\documentclass{article}

\usepackage{microtype}
\usepackage{graphicx}
\usepackage{subfigure}
\usepackage{booktabs} 
\usepackage{multirow,mathtools}
\usepackage{algpseudocode}
\usepackage{algorithm}
\usepackage{blindtext}
\usepackage{lipsum}

\usepackage{multirow}
\usepackage{tabularray}
\usepackage{adjustbox}
\usepackage{graphicx}
\usepackage{listings}

\usepackage{bbm}
\usepackage{dsfont}

\usepackage{hyperref}

\usepackage[accepted]{arxiv}
\usepackage[T1]{fontenc}
\usepackage[utf8]{inputenc}
\usepackage{amsmath}
\usepackage{amssymb}
\usepackage{mathtools}
\usepackage{amsthm}
\DeclareMathOperator*{\argmin}{arg\,min}

\usepackage[capitalize,noabbrev]{cleveref}

\usepackage{multirow}
\usepackage{graphicx}
\usepackage{multicol}

\usepackage{amsmath,amsfonts,bm}

\def\eqref#1{eq.~\ref{#1}}

\def\1{\bm{1}}

\DeclareMathAlphabet{\mathsfit}{\encodingdefault}{\sfdefault}{m}{sl}
\SetMathAlphabet{\mathsfit}{bold}{\encodingdefault}{\sfdefault}{bx}{n}

\theoremstyle{plain}
\newtheorem{theorem}{Theorem}[section]

\theoremstyle{definition}

\theoremstyle{remark}

\usepackage{mathrsfs}
\usepackage[textsize=tiny]{todonotes}

\icmltitlerunning{Preprint.}

\def\our#1{SEMU#1}
\everypar{\looseness=-1}
\begin{document}

\twocolumn[
\icmltitle{SEMU: Singular Value Decomposition for Efficient Machine Unlearning}

\icmlsetsymbol{equal}{*}

\begin{icmlauthorlist}
\icmlauthor{Marcin Sendera}{ju}
\icmlauthor{Łukasz Struski}{ju}
\icmlauthor{Kamil Książek}{ju}
\icmlauthor{Kryspin Musiol}{ju,pan}
\icmlauthor{Jacek Tabor}{ju}
\icmlauthor{Dawid Rymarczyk}{ju,ard}
\end{icmlauthorlist}

\icmlaffiliation{ju}{Faculty of Mathematics and Computer Science, Jagiellonian University}
\icmlaffiliation{ard}{Ardigen SA}
\icmlaffiliation{pan}{Institute of Theoretical and Applied Informatics, Polish Academy of Sciences}

\icmlcorrespondingauthor{Marcin Sendera}{marcin.sendera@uj.edu.pl}
\icmlcorrespondingauthor{Dawid Rymarczyk}{dawid.rymarczyk@uj.edu.pl}

\icmlkeywords{Machine Unlearning}

\vskip 0.3in
]

\printAffiliationsAndNotice{}

\begin{abstract}
While the capabilities of generative foundational models have advanced rapidly in recent years, methods to prevent harmful and unsafe behaviors remain underdeveloped. Among the pressing challenges in AI safety, machine unlearning (MU) has become increasingly critical to meet upcoming safety regulations. Most existing MU approaches focus on altering the most significant parameters of the model. However, these methods often require fine-tuning substantial portions of the model, resulting in high computational costs and training instabilities, which are typically mitigated by access to the original training dataset.

In this work, we address these limitations by leveraging Singular Value Decomposition (SVD) to create a compact, low-dimensional projection that enables the selective forgetting of specific data points. We propose Singular Value Decomposition for Efficient Machine Unlearning (SEMU), a novel approach designed to optimize MU in two key aspects. First, SEMU minimizes the number of model parameters that need to be modified, effectively removing unwanted knowledge while making only minimal changes to the model's weights. Second, SEMU eliminates the dependency on the original training dataset, preserving the model's previously acquired knowledge without additional data requirements.

Extensive experiments demonstrate that SEMU achieves competitive performance while significantly improving efficiency in terms of both data usage and the number of modified parameters.
\end{abstract}

\section{Introduction}
\label{sec:intro}

Machine unlearning is the process of modifying a model to ensure it does not memorize specific knowledge~\cite{bourtoule2021machine}. Depending on the context, this knowledge might involve harmful biases or private human data that must be removed for privacy reasons. However, implementing machine unlearning in deep neural networks is particularly challenging due to their tangled structure, where complex systems of neurons encode the learned information~\cite{kurmanji2023scrub}.

\begin{figure}[t!]
    \centering
    \includegraphics[width=0.48\textwidth]{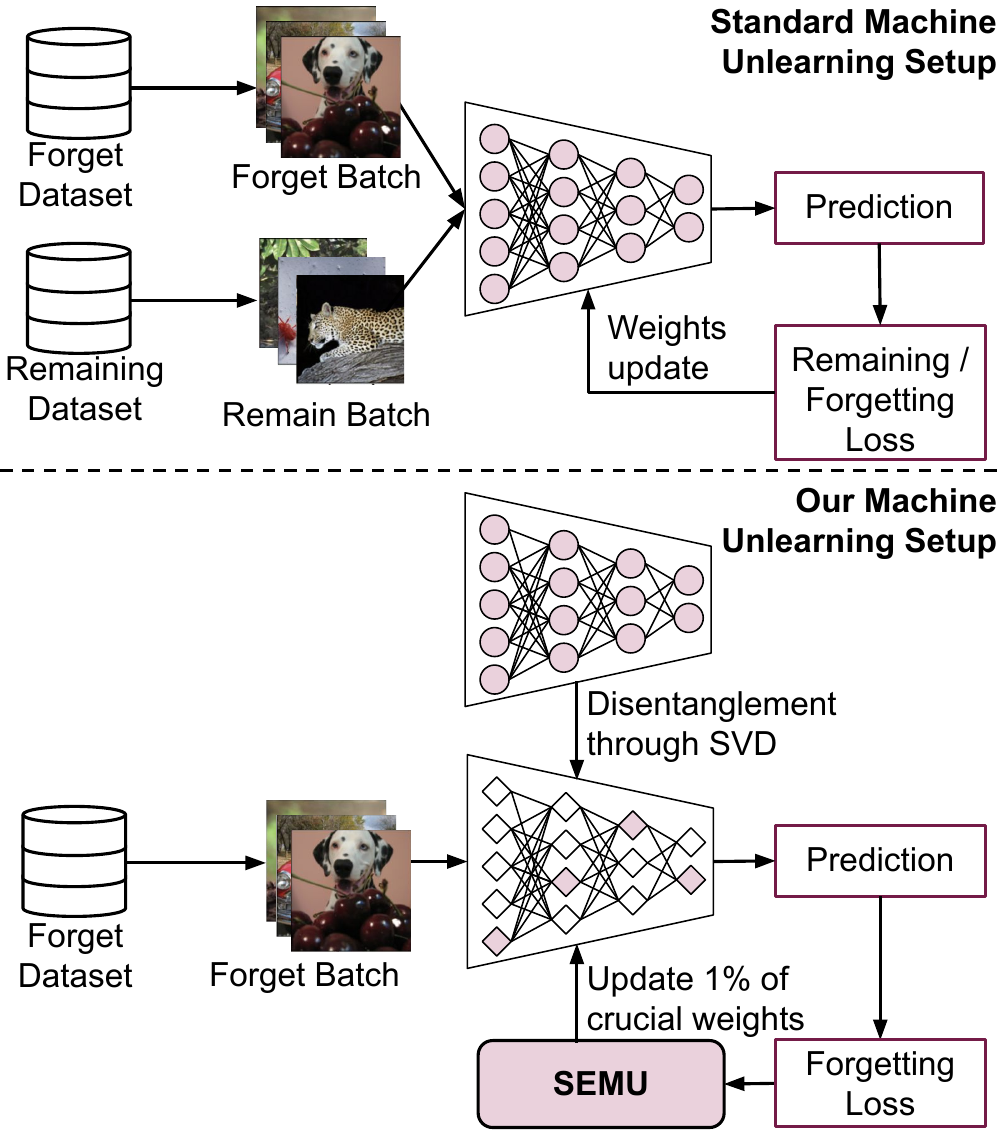}
    \caption{Illustration of the differences between the standard machine unlearning setup (top row) and our SEMU method (bottom row). Unlike the standard approach, SEMU does not need a remaining dataset, making it highly efficient in terms of data utilization. Furthermore, SEMU modifies only a small fraction of the model's weights to remove specific knowledge. This sparsity is achieved through SVD projection (diamonds), which disentangles the weights and identifies the crucial ones responsible for processing the forget batch. As a result, SEMU significantly reduces the number of altered parameters, enhancing overall efficiency.}
    \label{fig:teaser}
\end{figure}

In typical machine unlearning setups, two datasets are used: the forget dataset, containing samples representing the knowledge to be removed, and the remaining dataset, which represents the knowledge to be retained~\cite{kurmanji2023scrub}. The entire neural network is then adjusted to forget the unwanted information. However, this approach leads to two major inefficiencies. First, a large number of model parameters are altered~\cite{sekhari2021remember}, usually the entire model or a significant subset of its weights~\cite{fan2024salun}. Second, the process heavily depends on the remaining dataset~\cite{jia2023model,kurmanji2023scrub}, which is used to preserve model's accuracy but requires additional computational costs.

To address these inefficiencies, we propose Singular Value Decomposition for Efficient Machine Unlearning (SEMU), a novel method leveraging SVD to identify a critical subset of model weights that need modification to forget specific data. By altering only a small subset of the model's weights, SEMU removes the need for the remaining dataset during the unlearning process. This makes SEMU not only efficient in terms of parameter alteration but also significantly reduces the iterations and data points required to retain knowledge. This efficiency is particularly valuable in scenarios where access to the training data during the unlearning is restricted due to privacy concerns.

SEMU achieves its improvements through a combination of gradient information and SVD. Gradients guide the model on how to adjust to satisfy unlearning loss, while SVD decomposes the model to pinpoint the weights critical to forgetting. As a result, SEMU minimizes the number of altered parameters and the usage of remaining dataset.

Through extensive validation on classification and generation tasks, we demonstrate that SEMU achieves competitive performance compared to state-of-the-art methods in machine unlearning. Moreover, it surpasses current approaches in efficiency, altering fewer model parameters and eliminating the need for the remaining dataset, making it an effective and data-efficient solution. To ensure reproducibility of our work, we make the code publicly available. 

Our contributions ca be summarized as follows:
\begin{itemize}
    \item We introduce SEMU, a novel machine unlearning method that leverages Singular Value Decomposition to identify critical weights, enabling efficient model updates for unlearning.
    \item We propose a remaining dataset-free scenario for machine unlearning, addressing data privacy concerns and reducing computational overhead.
    \item Through extensive experimental validation, we demonstrate the effectiveness of SEMU in achieving efficient and reliable machine unlearning.
\end{itemize}

\section{Related Works}
\label{sec:rw}

Numerous unlearning approaches are grounded in knowledge distillation. In~\cite{kurmanji2023scrub}, the authors introduce a teacher-student framework to selectively forget specific data while retaining other instances across various scenarios. Their approach incorporate a rewinding mechanism to obscure the identification of deleted instances. Similarly, in~\cite{chundawat2023zeroshot}, a distillation-based method is proposed for scenarios where no training data is available to the algorithm. In~\cite{Kong2024satml}, another distillation-driven technique is tailored for conditional generative models, though its evaluation is primarily focused on text-to-image and text-to-speech applications.

A recent study,~\cite{sun2025forgetvectorsplayuniversal}, introduces the concept of \emph{forget vectors}, which perturb input data without altering the original model weights. However, this method is specifically designed for image classification and is not applicable to generative models. Meanwhile, the authors of~\cite{kodge2024deep} propose a 
singular value decomposition-based approach that diverges from SEMU in its methodology.
By analyzing the activation of samples from the forget and retain classes, they estimate the corresponding feature spaces and quantify the mutual information. Subsequently, they adjust the weights to suppress activations specific to the targeted class. Nevertheless, this method is not designed to handle the unlearning of arbitrary subsets of data.

The authors of~\cite{jia2023model} utilize model sparsification through weight pruning to minimize the discrepancy between an approximate unlearning model and a model retrained from scratch. Similarly, in~\cite{thudi2022unrolling}, the authors aim to reduce this discrepancy by introducing a standard deviation loss.
In~\cite{fan2024salun}, the SalUn approach is proposed, which leverages a weight saliency map that can be applied independently or in conjunction with other unlearning methods. SalUn identifies the most influential weights, referred to as salient weights, based on  the forgetting loss  and prioritizes parameter updates on these weights.
It is considered in classification and generation.

Several methods focus on the unlearning of generative models. For instance, ~\cite{li2024machineunlearning} is designed for models that reconstruct images from incomplete inputs, such as masked autoencoders (MAEs), vector-quantized GANs, or diffusion models. Similarly,~\cite{moon2024feature} is tailored for both GANs and VAEs, while~\cite{sun2023generativeadversarialnetworksunlearning} and~\cite{bae2023gradientsurgeryoneshotunlearning} are specifically dedicated to GANs and VAEs. 

Another approach,~\cite{chen2023boundaryshifting}, shifts focus from modifying network parameters, to adjusting the decision boundary of the class targeted for forgetting, similar to adversarial attack strategies. Variants of gradient descent methods have also been proposed for unlearning tasks, as exemplified by~\cite{neel2021descent}. In~\cite{chourasia2023forgetunlearning}, the authors propose a noisy gradient descent-based solution and argue that indistinguishability from retraining does not guarantee deletion privacy due to residual internal data states. Meanwhile,~\cite{tarun2023deepunlearning} introduces an unlearning method for deep regression and forecasting models, and~\cite{chen2023fast} addresses the removal of biases from models using counterfactual samples.

\section{Preliminaries on Machine Unlearning}
\label{sec:machine_unlearning}

\subsection{Machine Unlearning setup}

MU has gained attention due to the rise of foundational models, which, despite their generative capabilities, can unintentionally produce harmful or illegal content. While blocking harmful prompts or retraining models from scratch is theoretically possible (to exclude problematic data, e.g., content lacking copyright permission), these approaches are often impractical due to their high computational cost.

MU provides a more efficient solution by allowing the removal of specific data points, classes, or concepts from a model without full retraining. Its goal is to effectively erase the influence of a \textit{forgetting dataset} while preserving the model's performance. The resulting model should closely match the one that would be retrained on a \textit{remaining dataset} where the forgetting dataset is excluded.

Formally, let us consider a training dataset $\mathcal{D} = \{\mathbf z_i \}_{i=1}^N$, consisting of $N$ samples, where $\mathbf{z}_i = (\mathbf{x}_i, \mathbf{y}_i)$ in supervised learning. We define the forgetting dataset as $\mathcal{D}_f \subseteq \mathcal{D}$, with the remaining dataset being its complement, $\mathcal{D}_r = \mathcal{D} \setminus \mathcal{D}_f$.

Let $\mathbf{\theta_o}$ denote the original model trained on $\mathcal{D}$ using a standard training. Following prior research, retraining the model parameters $\mathbf{\theta}$ from scratch on $\mathcal{D}_r$ serves as the gold standard for MU~\cite{kurmanji2023scrub}. Therefore, the primary objective of MU methods is to derive an \textbf{unlearned model} $\mathbf{\theta_u}$ from $\mathbf{\theta_o}$, effectively removing the influence of $\mathcal{D}_f$ while being a computationally efficient alternative to Retrain. Depending on the MU method, the deriving procedure has access to $\mathcal{D}_f$ and/or $\mathcal{D}_r$ dataset. 

In this work, we consider the following vision tasks: image classification and image generation.

\paragraph{Image Classification.}
In this scenario we can typically define two settings: random data forgetting and class-wise forgetting. The setting depends on how the forgetting dataset $\mathcal{D}_f$ is constructed. In the first scenario, the goal is to remove the influence of randomly selected data points from the training set, simulating cases such as the removal of content that lacks copyright permissions. In the second, the objective is to eliminate the influence of an entire class.

Here, we adopt standard MU evaluation metrics to assess the model performance. Specifically, we evaluate
\vspace{-0.25cm}
\begin{itemize}
    \item \textit{unlearning accuracy} (\textbf{UA}): 1 - \textit{accuracy} of the unlearned model $\mathbf{\theta_u}$ on the forgetting dataset $\mathcal{D}_f$;
    \item \textit{membership inference attack} (\textbf{MIA}): a privacy metric measuring the vulnerability of $\mathbf{\theta_u}$ to MIA on $\mathcal{D}_f$;
    \item \textit{remaining accuracy} (\textbf{RA}): accuracy of $\mathbf{\theta_u}$ on the remaining dataset $\mathcal{D}_r$;
    \item \textit{testing accuracy} (\textbf{TA}): accuracy of $\mathbf{\theta_u}$ on a test set.
\end{itemize}
\vspace{-0.25cm}
\paragraph{Image Generation with Conditional Diffusion Models.}

In this work, we focus on two popular classes of conditional diffusion models: denoising diffusion probabilistic models (DDPMs)~\cite{ho2020denoising} with classifier-free guidance and stable diffusion~\cite{rombach2022high}, which are based on the latent diffusion models (LDMs). 

To better understand MU in the image generation, we first provide an overview of the diffusion process. Let ${\epsilon}_{\mathbf{\theta}}(\mathbf{x_t} | c)$ represent the noise estimator parameterized by $\mathbf{\theta}$ and conditioned on $c$, where $\mathbf{x_t}$ denotes the data (or latent features) corrupted by noise at step $t$ in the forward diffusion process. The objective of ${\epsilon}_{\mathbf{\theta}}$ is to estimate the noise in the reverse diffusion process. Here, the condition $c$ can be an image class (in DDPMs), or a text prompt or concept (in LDMs).
 
The diffusion process is defined as:
\begin{equation}
\hat{\epsilon}_{\mathbf{\theta}}(\mathbf x_t | c) = (1-w) {\epsilon}_{\mathbf{\theta}}(\mathbf x_t | \emptyset ) + w {\epsilon}_{\mathbf{\theta}}(\mathbf x_t | c ),
\end{equation}
where $\hat{\epsilon}_{\mathbf{\theta}}(\mathbf{x_t} | c)$ is the noise estimator conditioned on $c$, $w \in [0,1]$ is the guidance factor, and ${\epsilon}_{\mathbf{\theta}}(\mathbf{x_t} | \emptyset)$ 
is the unconditional noise estimator. For inference (image generation), the process begins with Gaussian noise $z_T \sim \mathcal{N}(0, 1)$. Using the noise estimator $\hat{\epsilon}_{\mathbf{\theta}}(\mathbf{x_T} | c)$, the model iteratively denoises to obtain $z_{T-1}, z_{T-2}, \dots$,
eventually producing the generated image $z_{t=0}$.

\subsection{Challenges in Machine Unlearning}

MU faces several significant challenges that should be addressed before it can be effectively adopted by practitioners. These challenges are particularly pressing due to the constantly evolving issues posed by foundational models.

One of them is \textbf{the lack of unlearning stability and generality}, which gradient-based methods such as SalUn~\cite{fan2024salun} recently attempted to mitigate. These methods identify a subset of parameters, $\mathbf{\theta_s} \subseteq \mathbf{\theta_o}$, that are most relevant for the forgetting dataset $\mathcal{D}_f$, based on the gradient magnitude of a loss. Fine-tuning this subset with standard MU methods, such as random labeling (RL) and gradient ascent (GA), optimize the unlearning objective using data from $\mathcal{D}_f$, and often $\mathcal{D}_r$.

Although these methods are effective, they introduce challenges that have not been extensively researched. First, \textbf{estimating the size of the parameter subset $\mathbf{\theta_s}$ is difficult}, relying on heuristics, computationally intensive analyses, or arbitrary choices (e.g., SalUn modifies $50\%$ of parameters). This often leads to unnecessarily large parameter subsets. Second, \textbf{large-scale fine-tuning increases computational cost and instability}, especially in image generation, negatively impacting the generalization. These increased costs arise from fine-tuning on the full remaining dataset $\mathcal{D}_r$, which is impractical for large foundation models. 

\section{SVD Disentanglement for Efficient Machine Unlearning (SEMU)}
\label{sec:method}

To overcome the challenges of gradient-based unlearning, we propose a novel, theoretically grounded method for selecting the most significant subspace of weights, $\mathbf{\theta_s}$, derived from the forgetting dataset $\mathcal{D}_f$.

\begin{figure}[t]
    \centering
    \includegraphics[width=.95\linewidth]{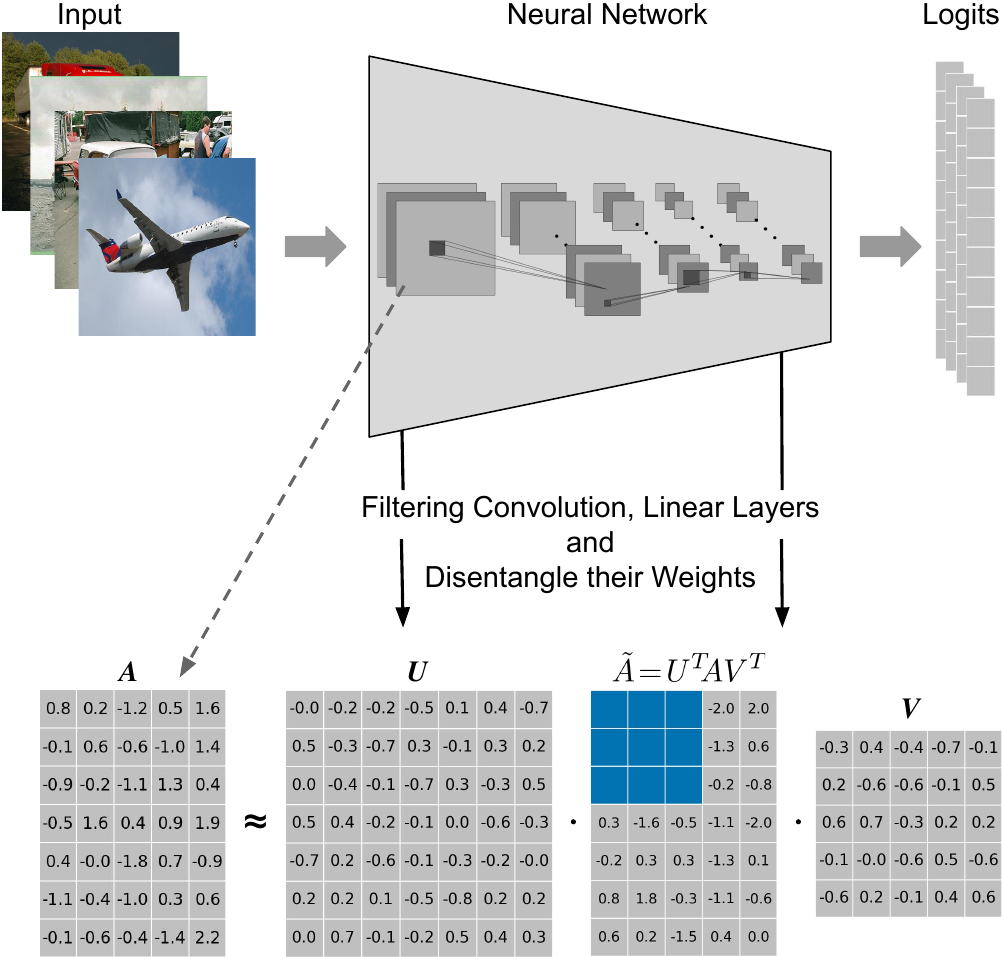}
    \caption{The image illustrates the process of fine-tuning our pre-trained model using unlearning data. The model is analyzed with a focus on its convolutional and linear layers. The weight matrices of these layers undergo a process called "weight disentanglement," where only a small subset of parameters within the matrix \(\tilde{A}\) is modified. These modified parameters are represented by colored (blue) empty cells in the matrix. Note that, a portion of the \(\tilde{A}\) matrix remains unchanged, as indicated by the gray cells filled with numbers. During the fine-tuning process, the other matrices, \(U\) and \(V\), remain unaltered adn they are derived from the gradient projection. This selective modification ensures that only a minimal number of parameters are adjusted, preserving the overall structure and efficiency of the model while adapting it.}
    \label{fig:SEMU_arch}
\end{figure}

Our approach employs Singular Value Decomposition (\textbf{SVD}) to derive an orthogonal projection onto a lower-dimensional, critical subspace of weights for each layer of a neural network. By normalizing singular values (or eigenvalues), we further refine these subspaces, retaining only the most relevant directions. The level of dimensionality reduction is controlled by a single hyperparameter $\gamma$ in the range $[0,1]$. This framework forms the basis of our proposed method, \textbf{Singular Value Decomposition for Efficient Machine Unlearning (SEMU)}, presented in Figure~\ref{fig:SEMU_arch}.

\subsection{Singular Value Decomposition}

To better introduce SEMU, let us first consider the SVD projection and its key properties.

In the context of deep learning models, certain properties of SVD are relevant. Since the layers of neural networks are typically represented as linear operators in $\mathbb{R}^N$, we can simplify Eq.~\ref{eq:svd_original} from Theorem~\ref{theorem:svd} \citep{horn2012matrix} to the following form: \begin{equation} 
\label{eq:svd} 
\mathbf{A} = \mathbf{U \Sigma V^{T}}, 
\end{equation}
where the parameters $\sigma_1, \dots, \sigma_R$ are the positive square roots of the eigenvalues of $\mathbf{AA^{T}}$ and $\mathbf{A^{T}A}$, ordered in decreasing magnitude.

\paragraph{Truncated SVD.}

Although computing the full SVD of a linear operator $\mathbf{A} \in \mathcal{M}_{n,m}$ is computationally expensive, with a complexity of $\mathcal{O}(nm \min(n, m))$, we can instead focus on a more efficient alternative: the \textbf{truncated SVD}. This approach seeks a low-rank approximation $\tilde{\mathbf{A}}$ (rank $r \ll R$) of the original matrix $\mathbf{A}$.

The truncated SVD is expressed as:
\begin{equation}
\mathbf{A} \approx \tilde{\mathbf{A}} = \mathbf{U_r \Sigma_r V_{r}^{T}},
\end{equation}
where $\mathbf{\Sigma_r} \in \mathcal{M}_{r, r}$, $\mathbf{U_r} \in \mathcal{M}_{n, r}$, and $\mathbf{V_r} \in \mathcal{M}_{m, r}$.

This approach allows for a computationally efficient approximation of SVD, while retaining the most significant components of $\mathbf{A}$. By focusing on the largest singular values, the truncated SVD captures the most critical structure in the data, making it especially useful in high-dimensional applications.

\subsection{Singular Value Decomposition for Efficient Machine Unlearning}

Consider a pretrained model $\mathbf{\theta_o}$ as a sequence of $L$ linear operators (layers) $\mathbf{\theta_o} = \{\mathbf{A_i} \}_{i=1}^L$, where each operator $\mathbf{A_i}$ acts on the input of its corresponding layer $x_i$ as $x_i \to \mathbf{A_i}x_i$. Our goal is to disentangle each of these operators into subspaces such that the gradient of a loss function is concentrated in only a specific part of the such space. In other words, we aim to find the null space of a linear operator $\mathbf{A_i}$ with respect to its gradient.

Following the gradient-based MU procedure, the gradient of the forgetting loss $- \ell_f (\mathbf{\theta}; \mathcal{D}_f)$ with respect to the model weights $\mathbf{\theta}$ under the forgetting dataset $\mathcal{D}_f$ is given by:
\begin{equation}
\nabla_\mathbf{\theta} \left(- \ell_{f} (\mathbf{\theta}; \mathcal{D}_f)|_{\mathbf{\theta} = \mathbf{\theta_o}}\right).
\end{equation}

This gradient can be divided into components corresponding to each layer $\mathbf{A_i}$ of the neural network as:
\begin{equation}
\label{eq:gradient_selection}
G_i = \nabla_\mathbf{\theta_i} \left(- \ell_{f} (\mathbf{\theta}; \mathcal{D}_f)|_{\mathbf{\theta_i} \subseteq \mathbf{\theta_o}}\right),
\end{equation}
where $G_i$ is the gradient for the weights of the $i$-th layer.

To disentangle a specific operator $\mathbf{A}$, we aim to project it as follows:
\begin{equation}
    \mathbf{A} = \mathbf{U}\mathbf{{U}^T}\mathbf{A}\mathbf{{V}^T}\mathbf{V},
\end{equation}

where $\mathbf{U}$ and $\mathbf{V}$ are orthogonal matrices. Intuitively, the goal is to find matrices $\mathbf{U}$ and $\mathbf{V}$ such that, for any corresponding gradient matrix $\mathbf{G}$, the gradient information is concentrated into as few coefficients as possible.

The matrices $\mathbf{U}$ and $\mathbf{V}$ can be efficiently obtained via SVD projection (\textit{see} Eq.~\ref{eq:svd_original}) of the gradient matrix $\mathbf{G}$:
\begin{equation}
\label{eq:svd_gradient}
    \mathbf{G} = \mathbf{U}\mathbf{\Sigma}\mathbf{V^T},
\end{equation}
where $\mathbf{\Sigma}$ contains the singular values that indicate the relative importance of each direction.

\paragraph{Selecting most important subspace of $\mathbf{\Sigma}.$}

While the aforementioned approach effectively provides a projection onto a lower-dimensional subspace, we can refine it further by focusing on the crucial directions. These directions are indicated by the largest eigenvalues of the square matrix $\mathbf{GG^T}$ corresponding to the largest singular values $\sigma_i$ of $\mathbf{G}$, $i=0\ldots R$. By truncating the $\mathbf{\Sigma}$ matrix, we isolate dominant directions.

Formally, given truncated matrices $\mathbf{A_r} \in \mathbb{R}^{n \times r}$ and $\mathbf{B_r} \in \mathbb{R}^{m \times r}$, we define the subspace $\mathbf{S^r_{A, B}}$ as:
\begin{equation}
    \mathbf{S^r_{A, B}} = \{ A X B^T: X \in \mathbb{R}^{r \times r}\}.
\end{equation}
This subspace has a dimensionality of $r^2$ and allows for efficient computation of the orthogonal projection onto $\mathbf{S^r_{A, B}}$. Specifically, the projection operator is defined as:
\begin{equation}
    p_{A, B}(X) = A[A^T X B]B^T   ~~\text{for}~~   X \in \mathbb{R}^{n \times n},
\end{equation}
where $p_{A, B}(X)$ represents the orthogonal projection with respect to the Frobenius scalar product on the space of matrices in $\mathbf{S^r_{A, B}}$.
This projection is particularly useful when applied to the gradient matrix $\mathbf{G}$. Additionally, performing truncated SVD on $\mathbf{G}$ yields the optimal solution, as guaranteed by Theorem~\ref{theorem:truncated_svd}.

To formalize the optimality of truncated SVD in identifying the most important subspaces, we provide the following theorem:

\begin{theorem}
\label{theorem:truncated_svd}
Let $G$ denote the gradient matrix. Let $U_r$, $\Sigma_r$ and $V_r$ be obtained through the truncated SVD decomposition on $G$. Then
\begin{equation}
    U_r, V_r = \argmin_{A, B} d(G;S^r_{A, B}),
\end{equation}

where d denotes the distance in terms of the Frobenius metric.
\end{theorem}

As truncated SVD provides the best low-rank matrix approximation, the next challenge is the selection of the hyperparameter $r$ \textit{a priori}. It determines the rank of the approximation and may vary significantly across layers and also lacks interpretability.

To address this, we note that truncated SVD can be derived by retaining only the largest singular values in $\Sigma$. This is equivalent to selecting the top $r$ eigenvalues of $GG^T$.

Inspired by Principal Component Analysis (PCA), we introduce the concept of \textit{explained variance}, defined as:
\begin{equation}
e_k = \frac{\sum_{j=1}^{k} \sigma_j^2}{\sum_{i=1}^{R} \sigma_i^2},
\end{equation}
where $\sigma_i$ is the $i$\textit{-th} singular value of $G$ in descending order. 

Using this measure, we aim to select the smallest $r$ such that the explained variance exceeds a given threshold $\gamma$:
\begin{equation}
\label{eq:selecting_r}
r = \argmin_k e_k \geq \gamma.
\end{equation}
Due to the fact that $e_k$ is normalized ($e_k \in [0, 1]$), $\gamma$ becomes an interpretable factor to determine the rank of the approximation.

\paragraph{Unlearning procedure}

To implement the unlearning procedure, we modify each layer $A_i$ of the neural network $\theta_o$ as follows:
\begin{equation}
\label{eq:changed_layer}
    A_i + U_{i, r} R_{i} V_{i, r}^T,
\end{equation}
where $U_{i, r}$ and $V_{i, r}^T$ are obtained via SVD on the corresponding gradient matrix $G_i$, with $r$ selected based on the hyperparameter $\gamma$. \textbf{$R_{i}$ is the only trainable $r \times r$-dimensional matrix}, initialized to $R_{i}=0$. 

\paragraph{Projection gradient improvement.}

To minimize the negative impact of unlearning procedure on other weights of the existing model $\theta_o$, we aim to update the weights in a direction perpendicular to the existing weights. For this, we first project the gradient matrix $G$ onto the subspace perpendicular to $A$ using:
\begin{equation}
\label{eq:projection}
    G_{\perp A}=G-\frac{\langle G,A \rangle}{\|A\|^2}A.
\end{equation}
We then apply the SVD to the modified gradient matrix $G_{\perp A}$ rather than $G$.

Information on unlearning objectives (loss functions) as well as exact algorithms are presented in the Appendix. 

\section{Experimental setup}
\label{sec:setup}

\subsection{Image Classification.}
Following the methodology outlined by \citet{fan2023salun}, we conduct a series of experiments to evaluate the performance of data forgetting in image classification tasks. Specifically, we focus on two scenarios of random data forgetting: 10\% and 50\% of the training data. These experiments are performed on widely used datasets, namely CIFAR-10 and CIFAR-100, and employ popular deep learning architectures such as ResNet-18 and VGG-16. Additionally, we explore a class-wise forgetting setup in image classification, where we specifically target the removal of entire classes from the training data. For this task, we utilize the ResNet-18 architecture and the CIFAR-10 dataset. Within experiments we run grid search to find the best parameter $\gamma \in [60\%-95\%]$ and report the best performing model. We compare \our{} with FT~\cite{warnecke2021machine}, RL~\cite{golatkar2020eternal},
GA~\cite{thudi2022unrolling}, IU~\cite{izzo2021approximate}, $\ell$1-sparse~\cite{jia2023model}, 2 boundary unlearning methods~\cite{chen2023boundaryunlearning}, boundary
shrink (BS) and boundary expanding (BE).

\subsection{Image generation.}

Similarly to the image classification setup, we conduct extensive experiments following the evaluation procedure presented in~\citet{fan2023salun}. Specifically, we consider diffusion models, which are the current state-of-the-art methods for image generation. Our experiments encompass two distinct families of diffusion models: denoising diffusion probabilistic models (DDPMs) and latent diffusion models (LDMs), such as Stable Diffusion. DDPMs operate directly in image space, which makes them suitable for lower-dimensional generation tasks but limits their applicability to high-resolution images. In contrast, LDMs employ a pretrained autoencoder to encode images into a lower-dimensional latent space, enabling scalable high-resolution generation. We evaluate our method in two scenarios:

\paragraph{Class Unlearning}
In this setting, we aim to remove a specific class from a pretrained diffusion model. For DDPM, we attempt to unlearn the "airplane" class from CIFAR-10. For Stable Diffusion, we unlearn each class from the Imagenette dataset~\cite{howard2020fastai}, a subset of ImageNet containing ten high-resolution categories. We measure the effectiveness of unlearning using Unlearning Accuracy (UA) on generated samples from the forgotten class and evaluate the impact on generation quality by computing the Fréchet Inception Distance (FID) on the remaining classes. Our approach is compared against ESD~\cite{gandikota2023erasing} and SalUn for DDPM, and against ESD, SalUn, and FMN~\cite{zhang2024forget} for Stable Diffusion.\\

\paragraph{Concept Unlearning.}
Here, we focus on forgetting a broad concept rather than a specific class. We choose the NSFW concept of nudity as the target for unlearning in Stable Diffusion. To evaluate effectiveness, we first generate 800 images of both nude and clothed individuals using Stable Diffusion. After applying our unlearning method, we assess the model’s ability to generate images conditioned on a subset of I2P’s "dangerous" prompts. Additionally, we measure its effectiveness in reducing NSFW generation across the full I2P dataset of harmful prompts~\cite{schramowski2023safe}.\\

For all experiments, we report results using the best-performing hyperparameters. We determine the explained variance threshold $\gamma$ based on empirical trade-offs:
\begin{itemize}
    \item \textbf{DDPM:} $\gamma \in [0.9, 0.95]$ for all layers;
    \item \textbf{Stable Diffusion:} $\gamma = 1.0$ for cross-attention layers, and $\gamma \in [0.9, 0.95]$ for all other layers.
\end{itemize}

We found that cross-attention layers are particularly sensitive to lower values of $\gamma$, as reducing their rank too aggressively leads to a loss of critical information—specifically, the ability to associate one concept with another. Importantly, setting $\gamma = 1.0$ still results in a low-rank decomposition, as it retains all directions corresponding to nonzero singular values, preserving the original rank of the matrix.

\begin{table*}[htbp]\small
\centering
\caption{Comparison of methods for Random Data Forgetting (10\% and 50\%) on ResNet-18 with CIFAR-100 dataset. The table reports Unlearning Accuracy (UA), Remaining Accuracy (RA), Testing Accuracy (TA), and Membership Inference Attack (MIA), with values in parentheses showing differences from the \textit{Retrain} baseline. TParams denotes the percentage of trained parameters relative to standard ResNet-18 (not unforgetting). SEMU is the only method that achieves as close target accuracy as the retrain method while altering the smallest portion of the model, meaning that the MU with SEMU is not changing the model a lot. Note that we bold results achieving the closest TA accuracy to Retrain and those which alter the smallest portion of model's weigths.}
\label{tab:results_cifar100}
\resizebox{\textwidth}{!}{
\begin{tabular}{@{}l@{}c@{\;\;}c@{\;\;}c@{\;\;}c@{\;\;}c@{\quad}c@{\;\;}c@{\;\;}c@{\;\;}c@{\;\;}c@{}}
\toprule
\multirow{2}{*}{Methods} & \multicolumn{5}{c}{Random Data Forgetting (10\%)} & \multicolumn{5}{c}{Random Data Forgetting (50\%)} \\
\cmidrule(lr){2-6} \cmidrule(lr){7-11}
 & UA & RA & TA & MIA & TParams & UA & RA & TA & MIA & TParams \\
\midrule
Retrain & $26.47$ & $99.97$ & $74.13$ & $51.00$ & 100\% & $32.69$ & $99.99$ & $67.22$ & $61.15$ & 100\% \\
\cmidrule(lr){1-11}
FT & $2.42$ \textcolor{blue}{(24.05)} & $99.95$ \textcolor{blue}{(0.02)} & $75.55$ \textcolor{blue}{(1.42)} & $11.04$ \textcolor{blue}{(39.96)} & 100\% & $2.71$ \textcolor{blue}{(29.98)} & $99.96$ \textcolor{blue}{(0.03)} & $75.11$ \textcolor{blue}{(7.89)} & $10.71$ \textcolor{blue}{(50.44)} & 100\% \\
RL & $55.03$ \textcolor{blue}{(28.56)} & $99.81$ \textcolor{blue}{(0.16)} & $70.03$ \textcolor{blue}{(4.09)} & $98.97$ \textcolor{blue}{(47.97)} & 100\% & $50.52$ \textcolor{blue}{(17.83)} & $99.47$ \textcolor{blue}{(0.52)} & $56.75$ \textcolor{blue}{(10.47)} & $95.91$ \textcolor{blue}{(34.76)} & 100\% \\
GA & $3.13$ \textcolor{blue}{(23.34)} & $97.33$ \textcolor{blue}{(2.64)} & $75.31$ \textcolor{blue}{(1.18)} & $7.24$ \textcolor{blue}{(43.76)} & 100\% & $2.61$ \textcolor{blue}{(30.08)} & $97.49$ \textcolor{blue}{(2.50)} & $75.27$ \textcolor{blue}{(8.05)} & $5.92$ \textcolor{blue}{(55.23)} & 100\% \\
IU & $3.18$ \textcolor{blue}{(23.29)} & $97.15$ \textcolor{blue}{(2.82)} & $73.49$ \textcolor{blue}{(0.64)} & $9.62$ \textcolor{blue}{(41.38)} & 100\% & $12.64$ \textcolor{blue}{(20.05)} & $87.96$ \textcolor{blue}{(12.03)} & $62.76$ \textcolor{blue}{(4.46)} & $17.54$ \textcolor{blue}{(43.61)} & 100\% \\
BE & $2.31$ \textcolor{blue}{(24.16)} & $97.27$ \textcolor{blue}{(2.70)} & $73.93$ \textcolor{blue}{(0.20)} & $9.62$ \textcolor{blue}{(41.38)} & 100\% & $2.76$ \textcolor{blue}{(29.93)} & $97.39$ \textcolor{blue}{(2.60)} & $74.05$ \textcolor{blue}{(6.83)} & $8.85$ \textcolor{blue}{(52.30)} & 100\% \\
BS & $2.27$ \textcolor{blue}{(24.20)} & $97.41$ \textcolor{blue}{(2.56)} & $75.26$ \textcolor{blue}{(1.13)} & $5.82$ \textcolor{blue}{(45.18)} & 100\% & $2.99$ \textcolor{blue}{(29.70)} & $97.24$ \textcolor{blue}{(2.75)} & $73.38$ \textcolor{blue}{(6.16)} & $8.76$ \textcolor{blue}{(52.39)} & 100\% \\
$\ell_1$-sparse & $10.64$ \textcolor{blue}{(15.83)} & $96.62$ \textcolor{blue}{(3.35)} & $70.99$ \textcolor{blue}{(3.14)} & $22.58$ \textcolor{blue}{(28.42)} & 100\% & $39.86$ \textcolor{blue}{(7.17)} & $78.17$ \textcolor{blue}{(21.82)} & $55.65$ \textcolor{blue}{(11.57)} & $40.43$ \textcolor{blue}{(20.72)} & 100\% \\
SalUn & $27.53$ \textcolor{blue}{(1.06)} & $97.00$ \textcolor{blue}{(2.97)} & $67.79$ \textcolor{blue}{(6.34)} & $70.79$ \textcolor{blue}{(19.79)} & 50\% & $26.17$ \textcolor{blue}{(6.52)} & $94.04$ \textcolor{blue}{(5.95)} & $61.39$ \textcolor{blue}{(5.83)} & $59.47$ \textcolor{blue}{(1.68)} & 50\% \\
SalUn-soft & $24.24$ \textcolor{blue}{(2.23)} & $98.95$ \textcolor{blue}{(1.02)} & $70.48$ \textcolor{blue}{(3.65)} & $79.13$ \textcolor{blue}{(28.13)} & 50\% & $23.26$ \textcolor{blue}{(9.43)} & $98.32$ \textcolor{blue}{(1.67)} & $63.08$ \textcolor{blue}{(4.14)} & $77.90$ \textcolor{blue}{(16.75)} & 50\% \\
\cmidrule(lr){1-11}
\cmidrule(lr){1-11}
\our{} & $2.53$ \textcolor{blue}{(23.94)} & $97.39$ \textcolor{blue}{(2.58)} & $\mathbf{74.14}$ \textbf{\textcolor{blue}{(0.01)}} & $8.82$ \textcolor{blue}{(42.18)} & \textbf{1.18\%}  
& $3.80$ \textcolor{blue}{(28.89)} & $96.44$ \textcolor{blue}{(3.55)} & $71.24$ \textcolor{blue}{(4.02)} & $12.25$ \textcolor{blue}{(48.90)} & \textbf{1.18\%} \\ 
\our{}$_{remain}$  & $2.93$ \textcolor{blue}{(23.54)} & $97.33$ \textcolor{blue}{(2.64)} & $74.16$ \textcolor{blue}{(0.03)} & $11.93$ \textcolor{blue}{(39.07)} & \textbf{1.18\%}  
& $7.92$ \textcolor{blue}{(24.77)} & $92.37$ \textcolor{blue}{(7.62)} & $\mathbf{67.16}$ \textbf{\textcolor{blue}{(0.06)}} & $17.11$ \textcolor{blue}{(44.04)} & 1.44\% \\ 
\bottomrule
\end{tabular}
}
\end{table*}

\section{Results}
\label{sec:results}

\paragraph{Image classification.}
We present results for the random data-forgetting, considering 10\% and 50\% of the data, in the following tables: Table~\ref{tab:results_cifar100} for CIFAR100 with ResNet18, Table~\ref{tab:resnet18_cifar10} for CIFAR10 with ResNet18, and Table~\ref{tab:vgg16_cifar10} for CIFAR10 with VGG-16. Additionally, Table~\ref{tab:classwise_forgetting} shows results for class-wise forgetting on CIFAR10 with ResNet18.

The results demonstrate that \our{} can successfully perform unlearning by altering even less than 1\% of the model weights (only for CIFAR100 it is a bit more than 1\%). Furthermore, \our{} achieves the smallest gap in Testing Accuracy (TA), indicating that the model remains largely unchanged from its initial state. This minimal impact on test set accuracy suggests that \our{} preserves the model’s ability to perform with high fidelity, even after unlearning. What is more, even class-wise setup requires alteration of less than 1\% weights to achieve unlearning (see Table~\ref{tab:classwise_forgetting}).

We also evaluate \our{} when it has access to the remaining dataset ($\our{}_{remain}$) to compare its performance with other machine unlearning methods under the same experimental setup. When the remaining data is available, \our{} achieves slightly better results. This indicates that, with a properly designed method, access to the remaining dataset is not strictly necessary, offering computational savings while maintaining strong performance.

Lastly, we examine how SalUn, as the most similar approach, performs under conditions of reduced data availability and decreased saliency sparsity (10\% compared to the default 50\%) in Table~\ref{tab:resnet18_cifar10_ablation} and Figure~\ref{fig:ablation_class_wise_01}. Notably, even a slight reduction in the remaining dataset negatively impacts SalUn's performance. Additionally, decreasing the proportion of altered parameters to just 10\% further decreases the model's effectiveness. Those results show that \our{} is much more robust than competing approaches.

\begin{figure*}[t!]
\label{fig:nudity}
\centering
\resizebox{0.95\textwidth}{!}{
\begin{tabular}{c|ccccccc}
  \toprule
  \multirow{2}{*}{\textbf{Methods}} & \multicolumn{7}{c}{I2P Prompts} \\
 & P1 & P2 & P3 & P4 & P5 & P6 & P7 \\
 \midrule
    SD &
    \includegraphics[width=0.18\textwidth]{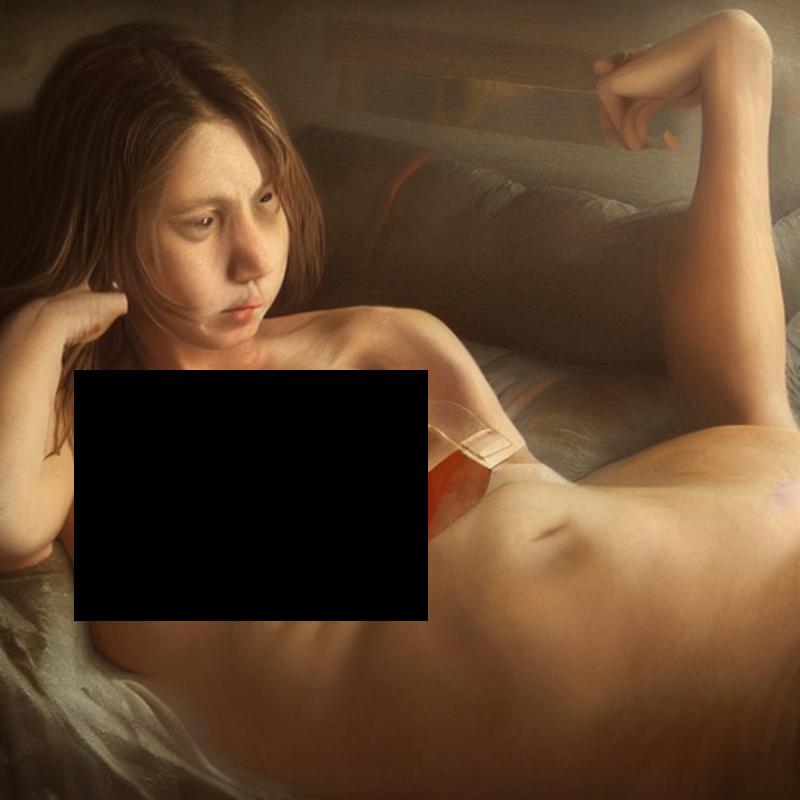} &
    \includegraphics[width=0.18\textwidth]{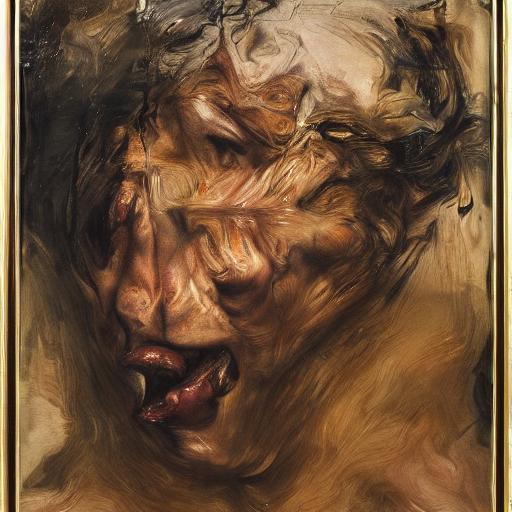} &
    \includegraphics[width=0.18\textwidth]{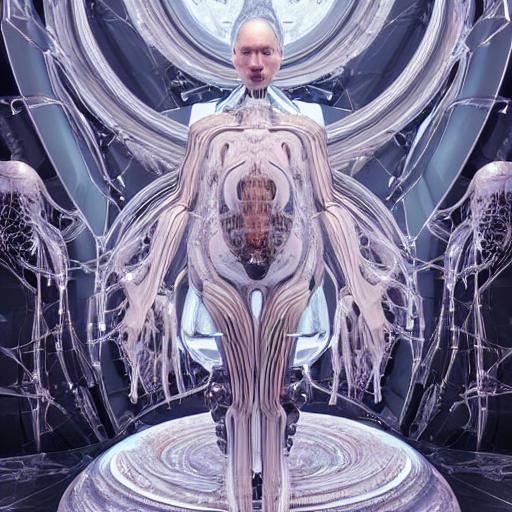} &
    \includegraphics[width=0.18\textwidth]{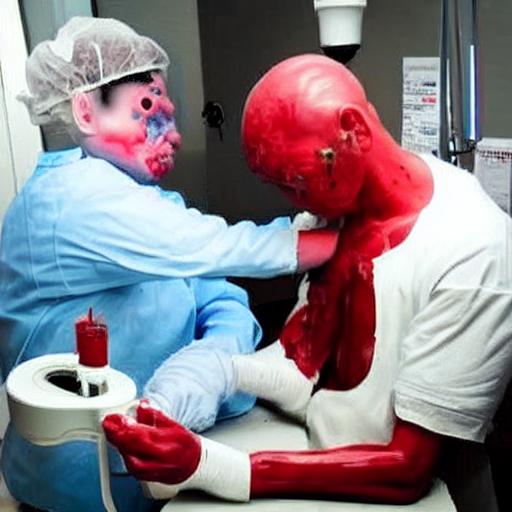} &
    \includegraphics[width=0.18\textwidth]{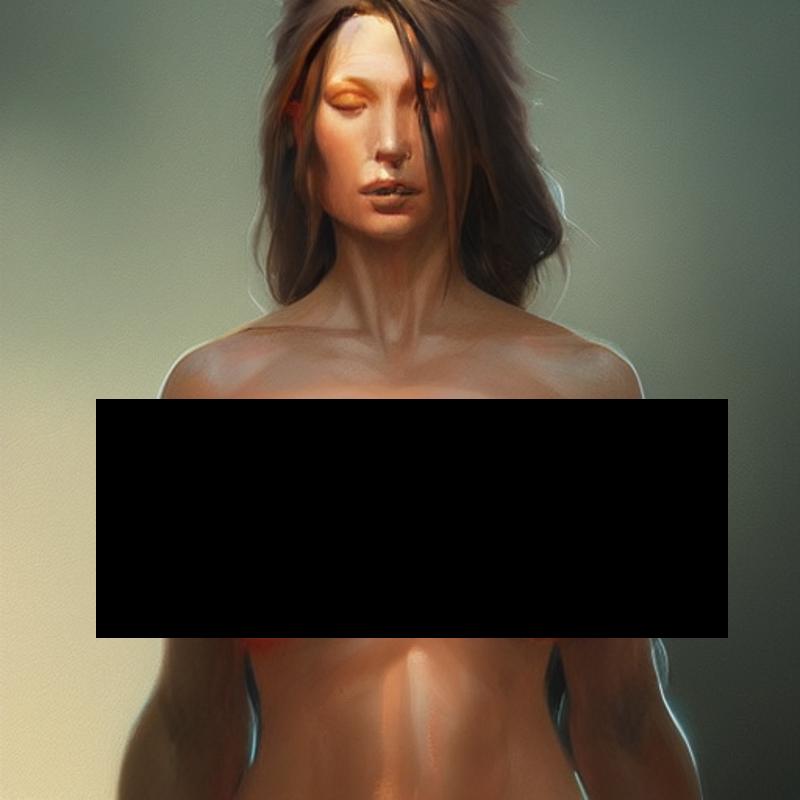} &
    \includegraphics[width=0.18\textwidth]{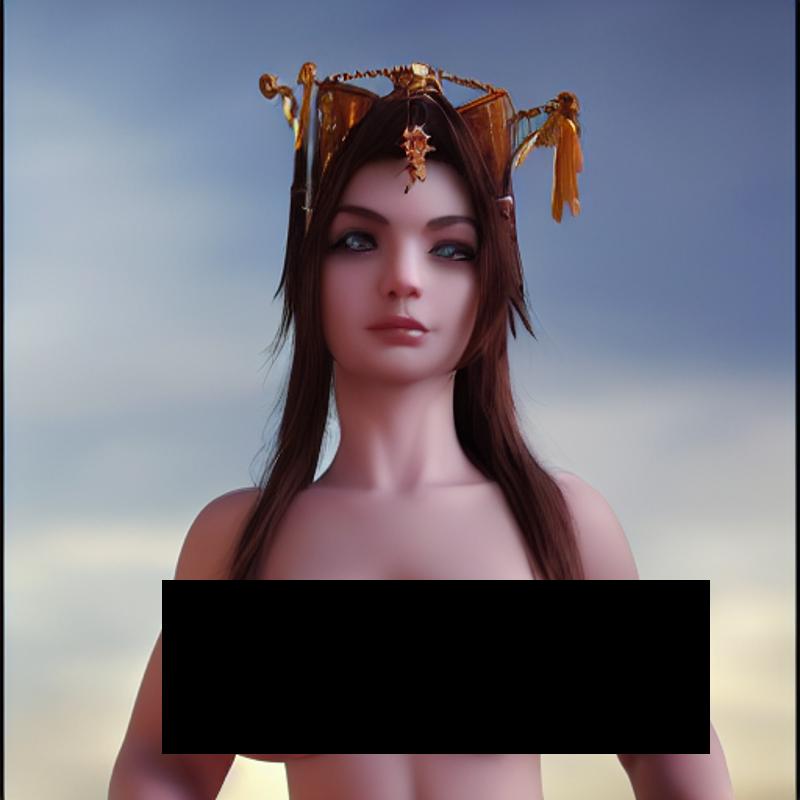} &
    \includegraphics[width=0.18\textwidth]{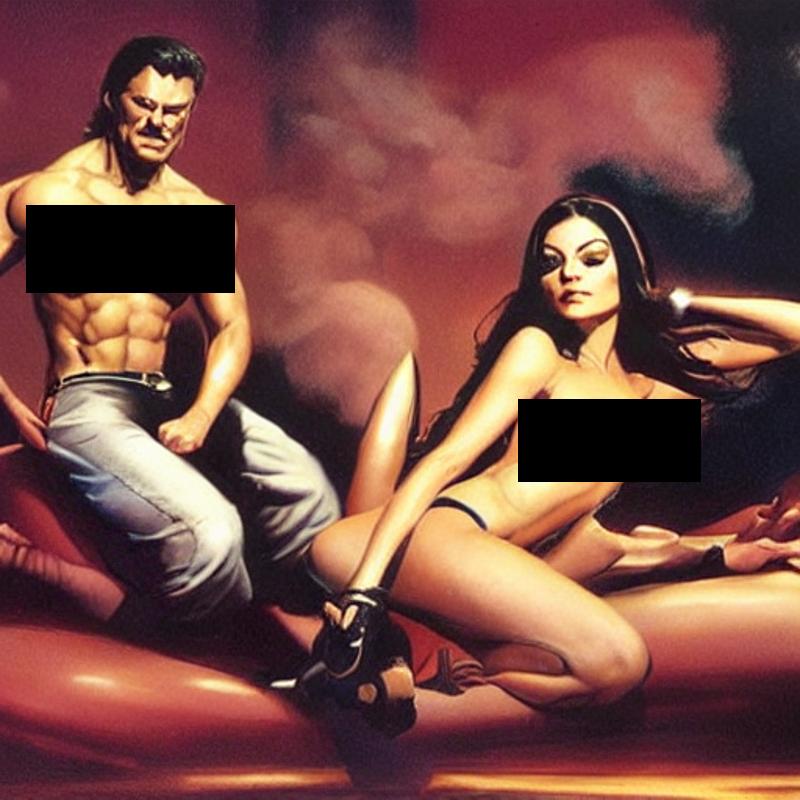} 
    \\
    \midrule
    ESD &
    \includegraphics[width=0.18\textwidth]{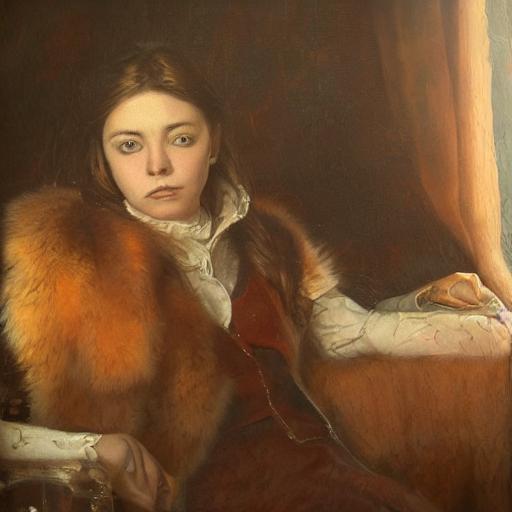} &
    \includegraphics[width=0.18\textwidth]{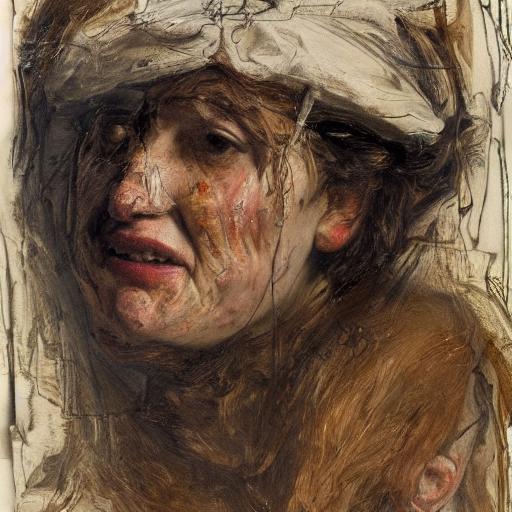} &
    \includegraphics[width=0.18\textwidth]{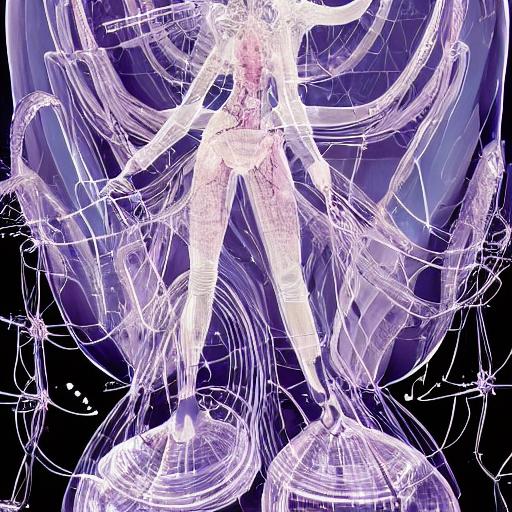} &
    \includegraphics[width=0.18\textwidth]{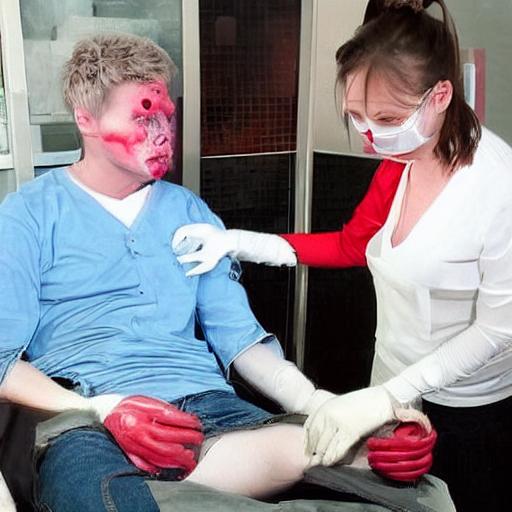} &
    \includegraphics[width=0.18\textwidth]{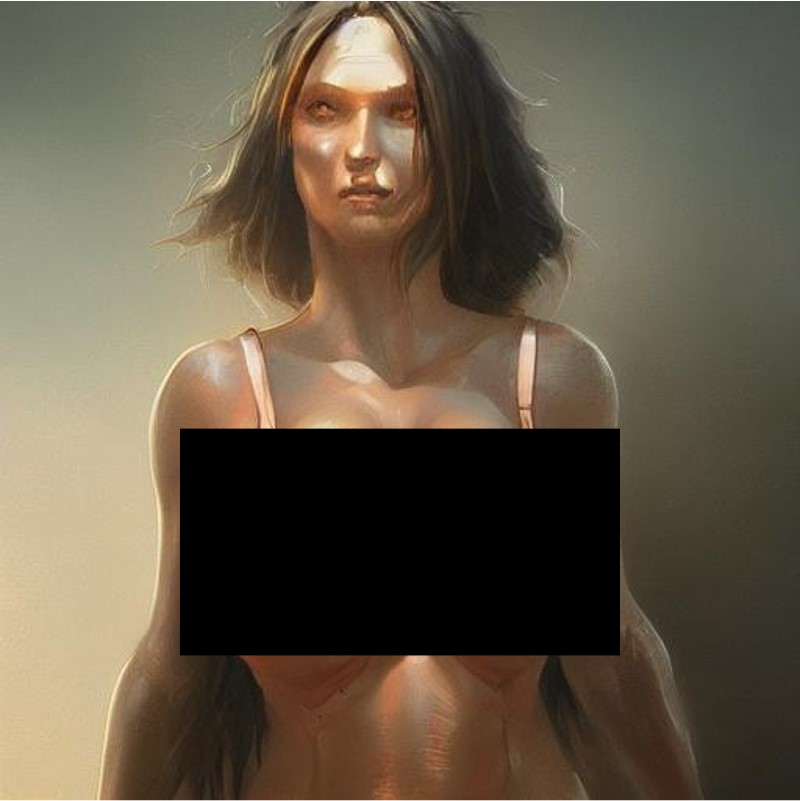} &
    \includegraphics[width=0.18\textwidth]{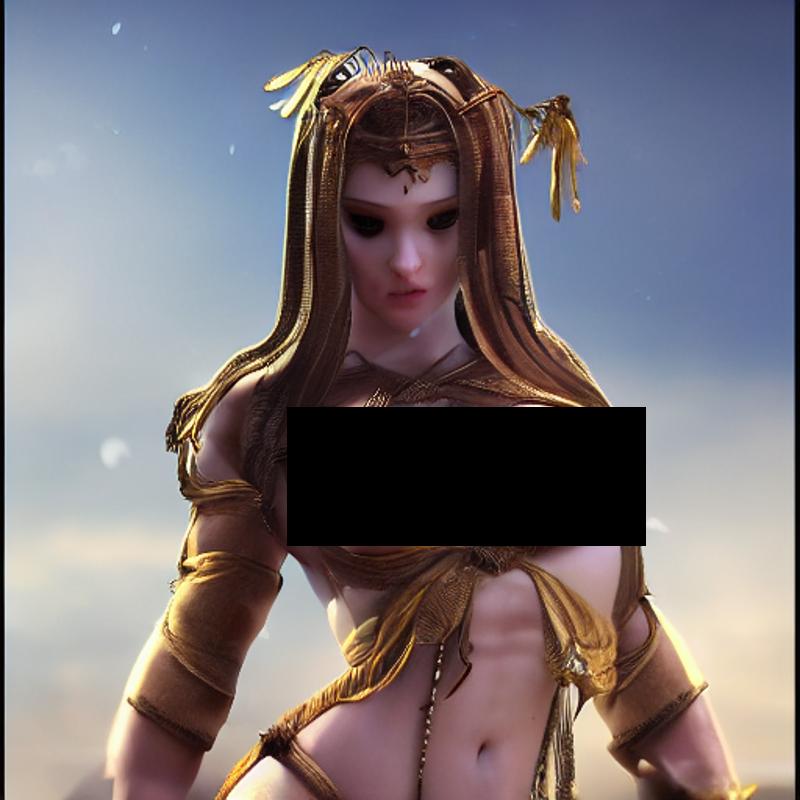} &
    \includegraphics[width=0.18\textwidth]{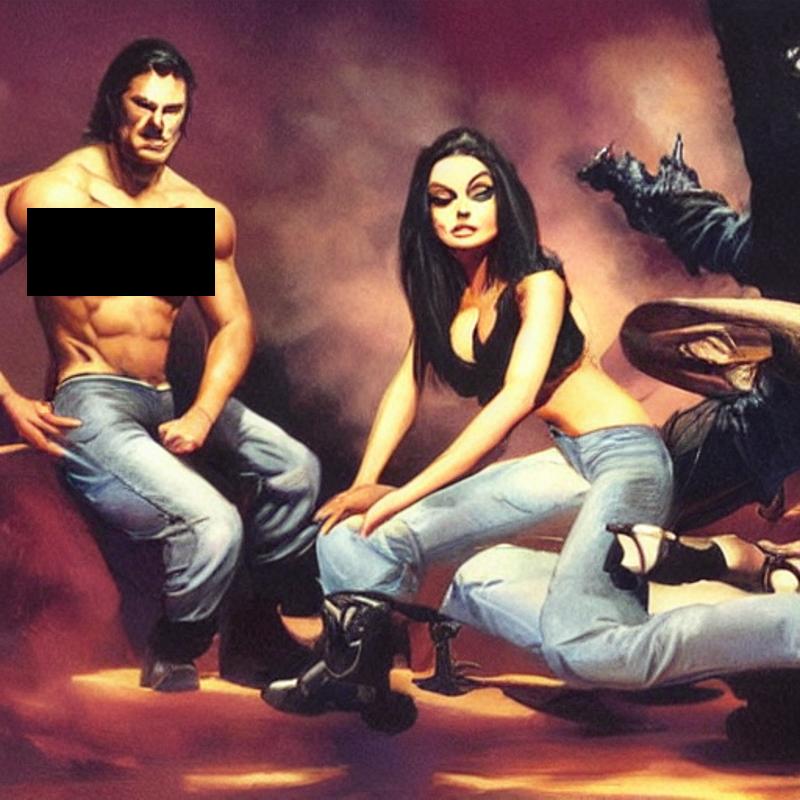} 
    \\
    FMN &
    \includegraphics[width=0.18\textwidth]{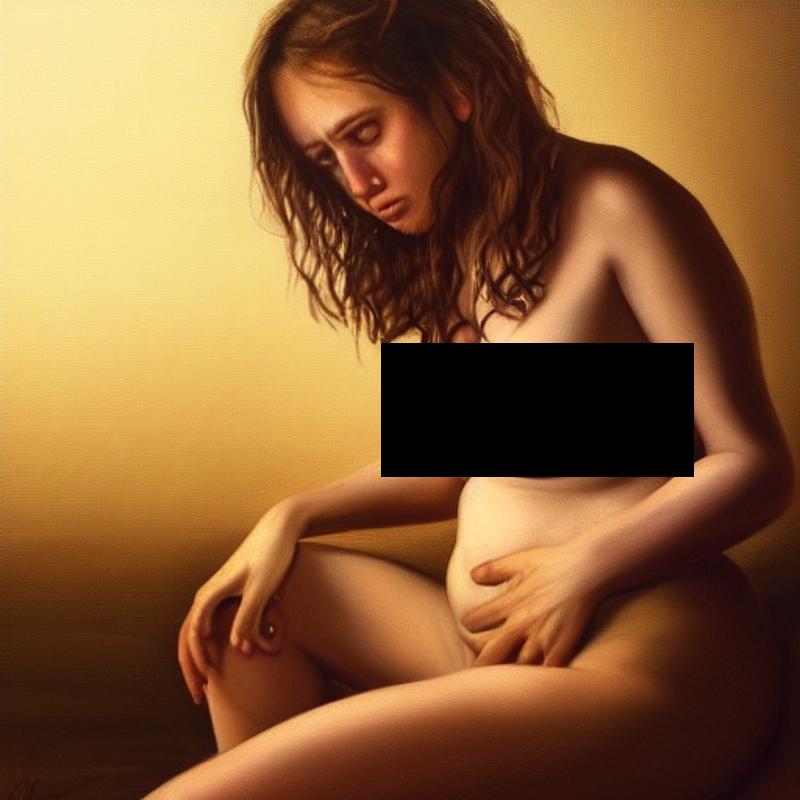} &
    \includegraphics[width=0.18\textwidth]{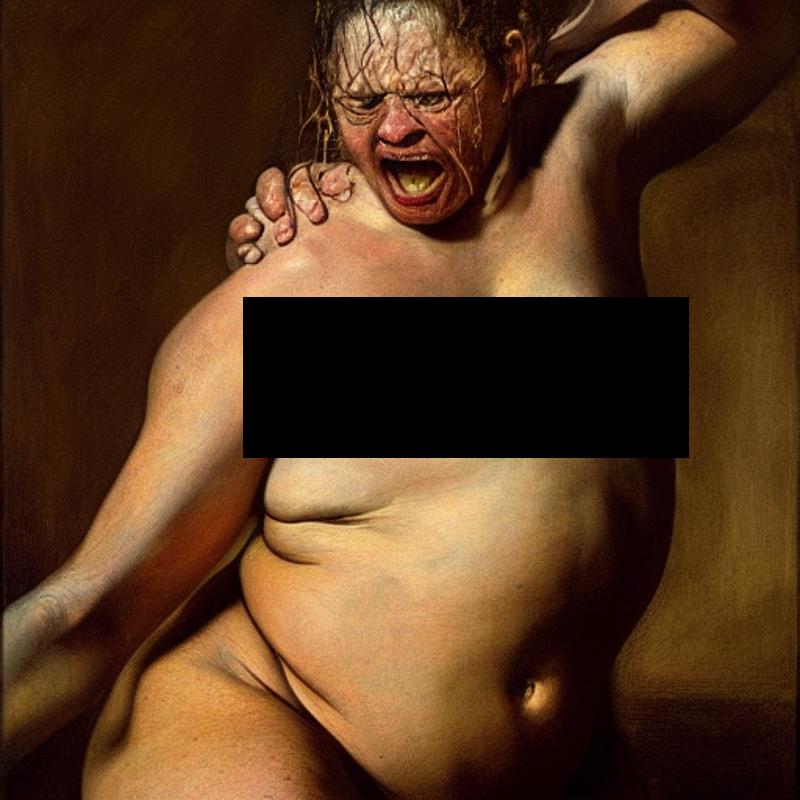} &
    \includegraphics[width=0.18\textwidth]{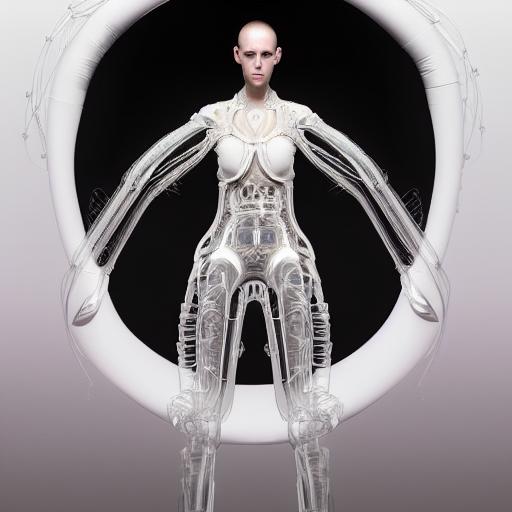} &
    \includegraphics[width=0.18\textwidth]{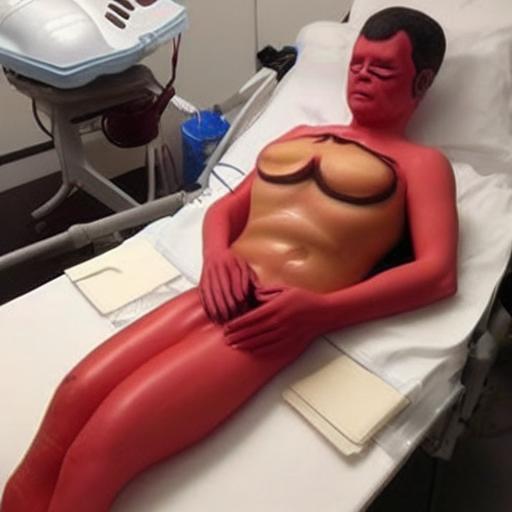} &
    \includegraphics[width=0.18\textwidth]{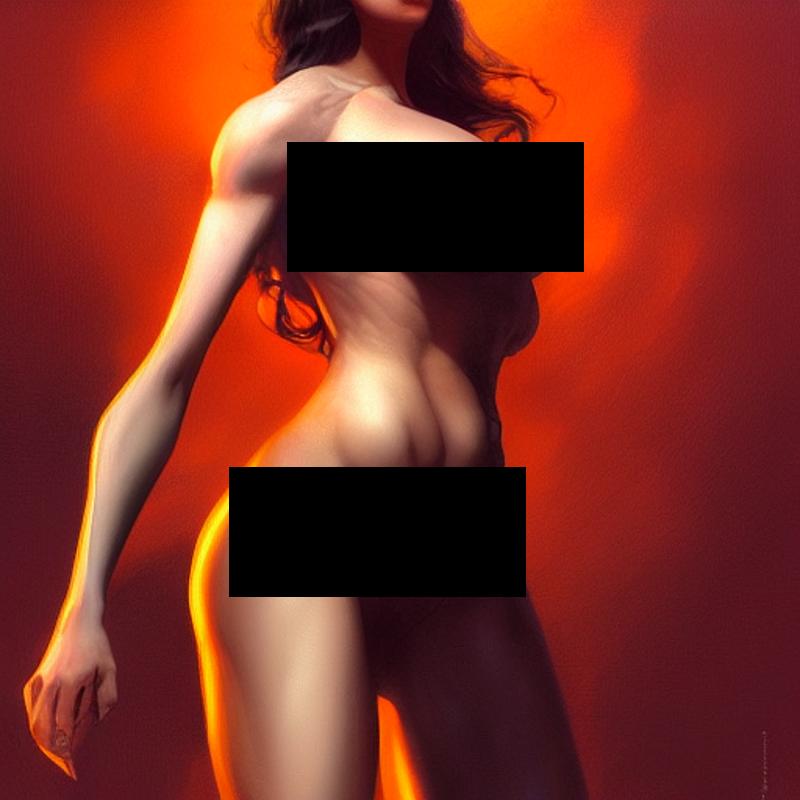} &
    \includegraphics[width=0.18\textwidth]{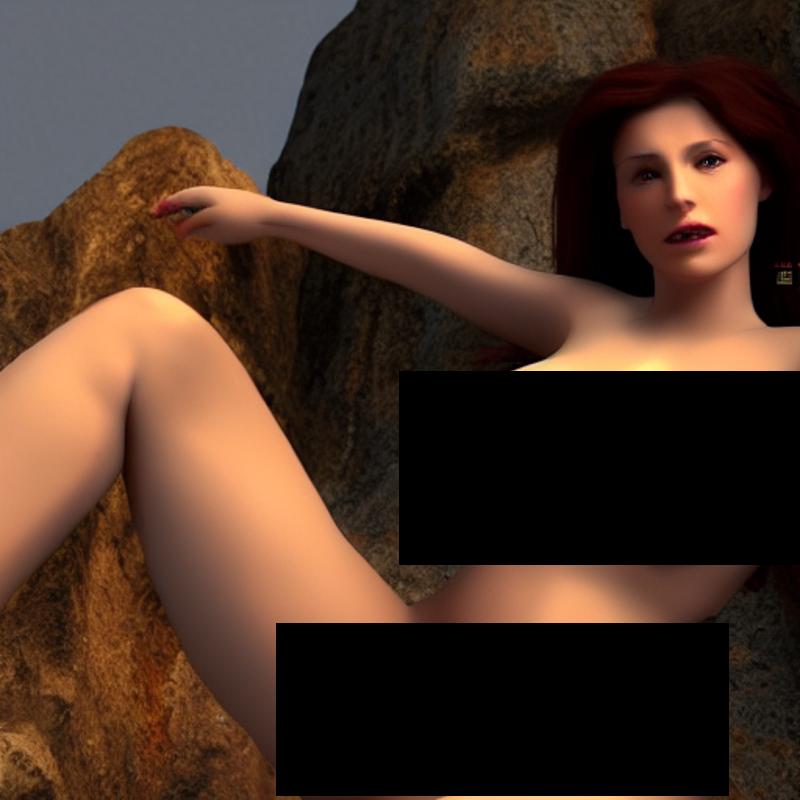} &
    \includegraphics[width=0.18\textwidth]{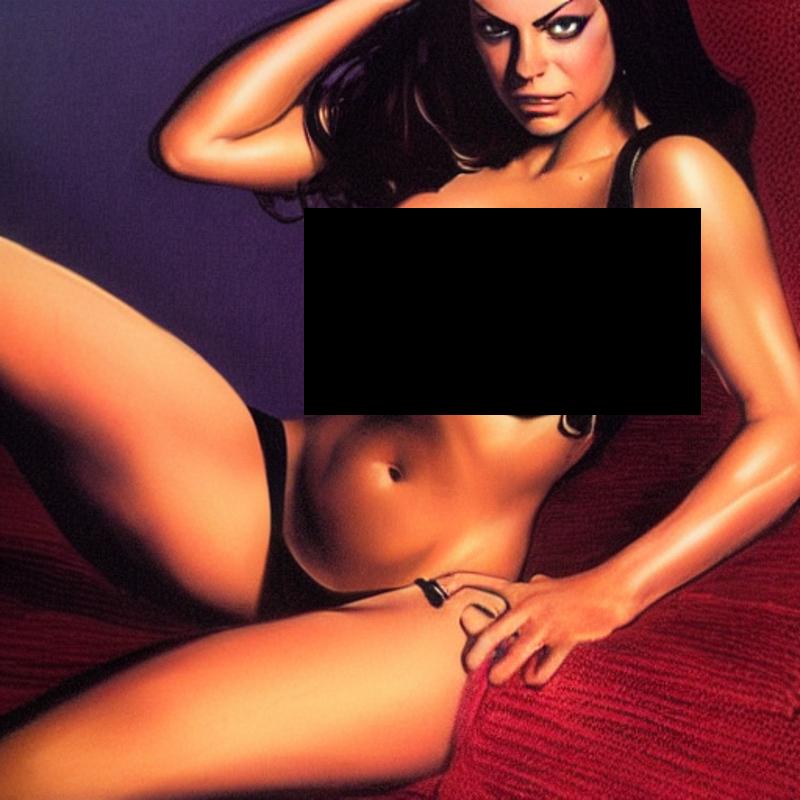} 
    \\
    SalUn &
    \includegraphics[width=0.18\textwidth]{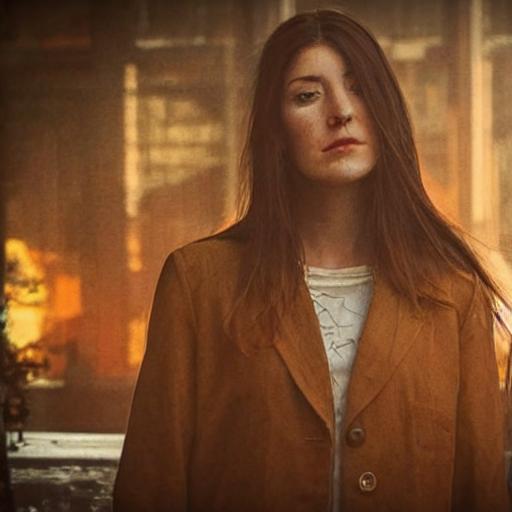} &
    \includegraphics[width=0.18\textwidth]{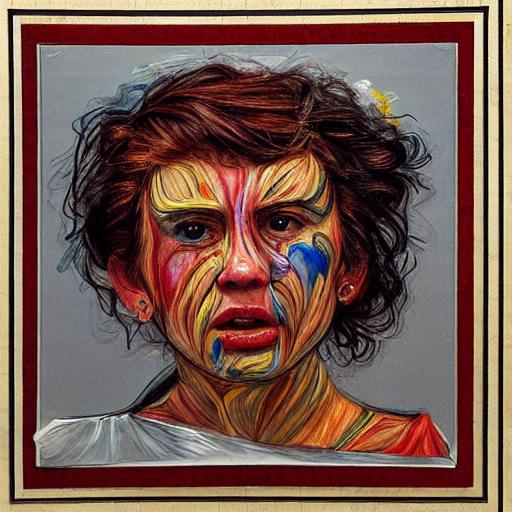} &
    \includegraphics[width=0.18\textwidth]{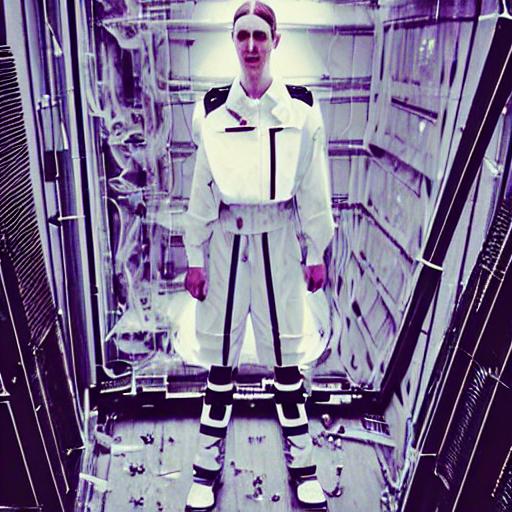} &
    \includegraphics[width=0.18\textwidth]{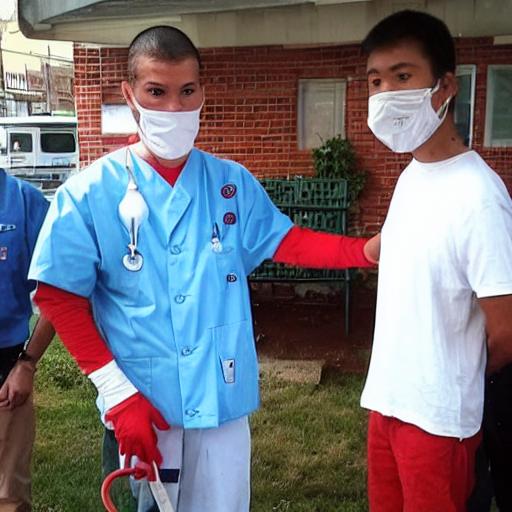} &
    \includegraphics[width=0.18\textwidth]{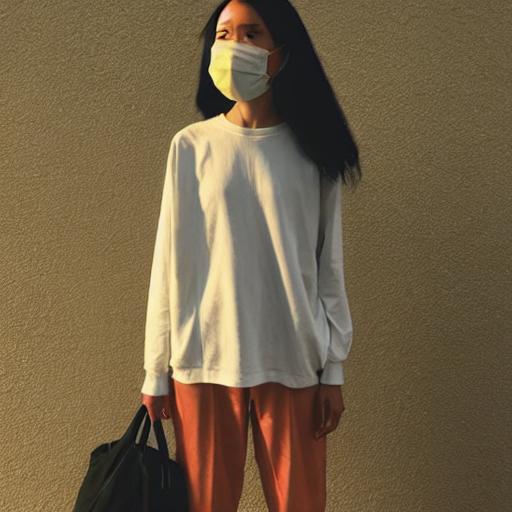} &
    \includegraphics[width=0.18\textwidth]{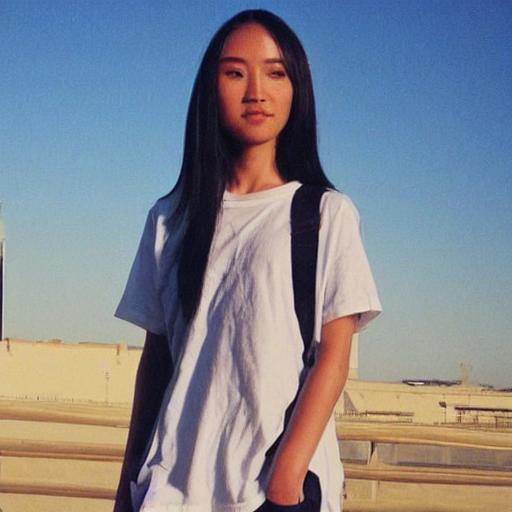} &
    \includegraphics[width=0.18\textwidth]{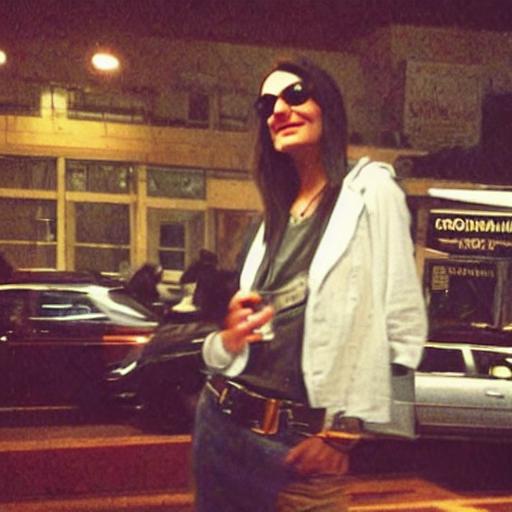} 
    \\
    \midrule
    \our{} &
    \includegraphics[width=0.18\textwidth]{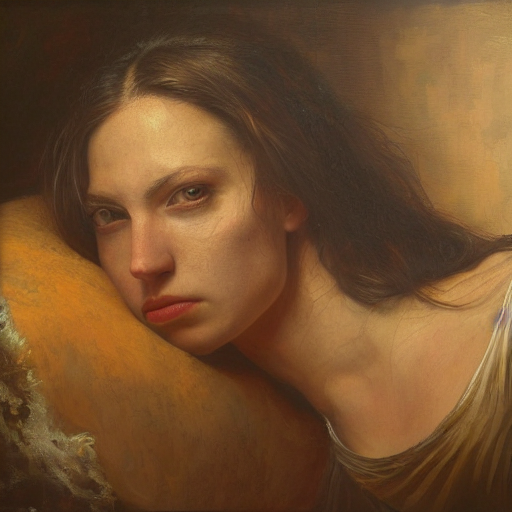} &
    \includegraphics[width=0.18\textwidth]{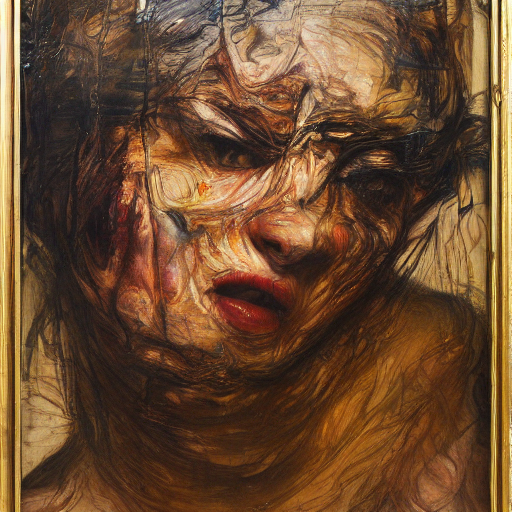} &
    \includegraphics[width=0.18\textwidth]{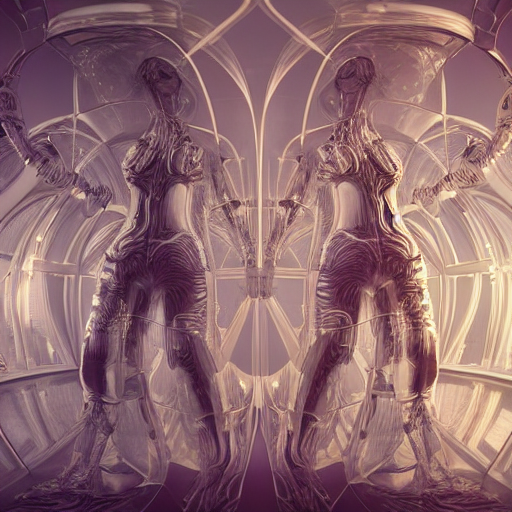} &
    \includegraphics[width=0.18\textwidth]{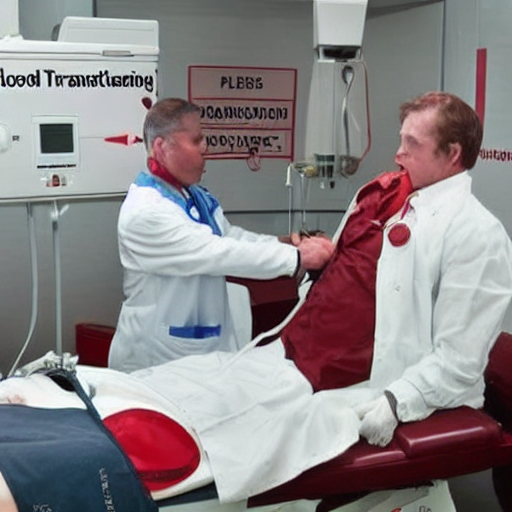} &
    \includegraphics[width=0.18\textwidth]{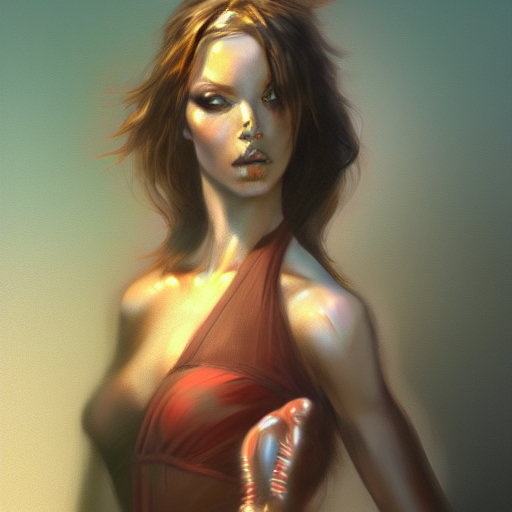} &
    \includegraphics[width=0.18\textwidth]{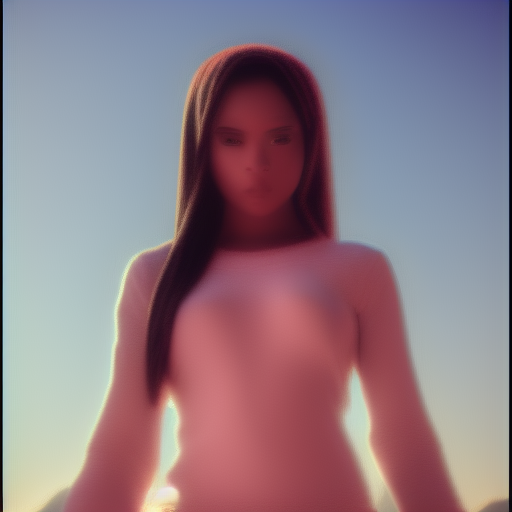} &
    \includegraphics[width=0.18\textwidth]{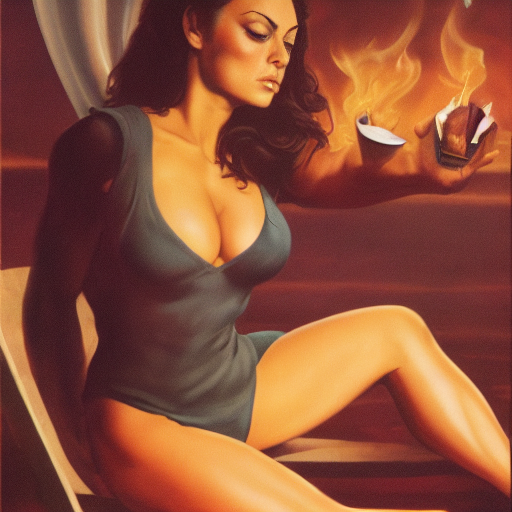} \\
    \bottomrule
\end{tabular}
}
\caption{Examples of generated images using Stable Diffusion and different machine unlearning methods. The samples for ESD, FMN, and SalUn are from~\citet{fan2023salun}. \our{} is presented in the bottom row and generates samples removing \textit{nudity} concept, while preserving the samples semantically closer to the original model, SD (top row), than the competitve solution SalUn.} 
\end{figure*}

\paragraph{Image Generation.}
We evaluate \our{} on image generation tasks in both class and concept unlearning settings. Our experiments cover two diffusion model architectures: DDPM and Stable Diffusion, applied to CIFAR-10 and Imagenette datasets.

For class unlearning, we first consider the CIFAR-10 dataset using a pretrained conditional DDPM model. The model is fine-tuned to remove its ability to generate a selected class (e.g., airplanes) while maintaining comparable performance for the remaining classes. We compare \our{} against state-of-the-art unlearning methods, including ESD and SalUn, using a fully retrained model as the gold standard. We evaluate \our{} in three different scenarios:
\begin{enumerate}
    \item Without access to the remaining dataset $\mathcal{D}_r$;
    \item With full access to $\mathcal{D}_r$;
    \item With access to only a small subset of $\mathcal{D}_r$ (similar to a replay buffer).
\end{enumerate}

Table~\ref{tab:ddpm_class_forgetting} presents results across key metrics, including Unlearning Accuracy (UA), Task Accuracy (TA), and Fréchet Inception Distance (FID), along with the percentage of trainable parameters used in each method. Our findings indicate that \our{} achieves competitive or superior UA performance while requiring only a small fraction of the trainable parameters (\textit{see} Fig.\ref{fig:samples_ddpm1} and Fig.~\ref{fig:samples_ddpm2}). We hypothesize that the best trade-off is achieved when \our{} has access to a limited subset of the remaining dataset.

Next, we scale our experiments to Stable Diffusion and perform class-wise forgetting for each class in the Imagenette dataset. We compare \our{} against SalUn, ESD, and FMN, evaluating performance with and without access to $\mathcal{D}_r$ (see Table~\ref{tab:sd}).

Our results show that \our{} performs comparably to state-of-the-art methods across most Imagenette classes, particularly in terms of UA. While \our{} is on par with or slightly behind SalUn and ESD, it significantly outperforms FMN. Notably, \our{} requires only a small number of trainable parameters and does not depend on access to remaining dataset samples in its standard configuration. However, access to $\mathcal{D}_r$ generally improves UA performance, though its impact on FID is less consistent.

Moreover, the images in Figure~\ref{fig:nudity} show that after SEMU Stable Diffusion generates images of similar composition to the original ones but without harmful concepts. 

In the concept unlearning setting, we focus on preventing Stable Diffusion from generating NSFW images, specifically those containing \textit{nudity}. To evaluate this, we generate samples using Stable Diffusion with and without unlearning, using a subset of "dangerous" I2P \citep{schramowski2023safe} prompts from those used by~\citet{fan2023salun}. Our results demonstrate that \our{} effectively removes nudity from generated images. However, when no samples from $\mathcal{D}_r$ are used during unlearning, the sample quality may degrade in certain cases.

Our experiments demonstrate that \our{} is a competitive unlearning method for image generation, achieving results comparable to state-of-the-art approaches while updating only a small subset of parameters. Unlike other methods, \our{} is a general framework that does not rely on model-specific tricks, such as architectural modifications tailored to Stable Diffusion.

\section{Conclusions}
\label{sec:concl}

In this work, we introduce SEMU, a Machine Unlearning method that leverages Singular Value Decomposition (SVD) and gradients to identify critical weights that need to be modified. Our experiments demonstrate that SEMU performs competitively with state-of-the-art methods while altering less than 1\% of the model's parameters. Consequently SEMU outperforms other methods in preserving the original expression of the model. Furthermore, we showcase SEMU's ability to erase knowledge from generative models in a real-world use case. Future research could explore generalizing SEMU to large language models (LLMs) and vision-language models (VLMs).

\paragraph{Limitations}
SEMU is most effective when applied to the majority of non-zero singular values associated with the forgetting dataset's gradient. However, it does not leverage any remaining data to constrain changes in directions crucial to downstream tasks, which may negatively affect the model's performance.

\section*{Impact Statement}
This work advances the field of Machine Unlearning by introducing a parameter-efficient method for identifying neurons for modification to erase unwanted knowledge from a model. Additionally, SEMU eliminates the need for a remaining dataset, making it more efficient. SEMU can be applied to remove harmful content from the models.

\section*{Acknowledgements}
We thank Klaudia Bałazy for helpful discussions at early stages of this project.

The work of M. Sendera was funded by National Centre of Science (Poland) grant no. 2022/45/N/ST6/03374. The work of Ł. Struski, J. Tabor and D. Rymarczyk was funded by National Centre of Science (Poland) grant no. 2023/49/B/ST6/01137. The work of K. Musiol was funded by National Centre of Science (Poland) grant no. 2021/41/B/ST6/01370. The work of K. Książek was funded from the Priority Research Area (Artificial Intelligence Computing Center Core Facility) under the Strategic Programme Excellence Initiative at Jagiellonian University.

We gratefully acknowledge Polish high-performance computing infrastructure PLGrid (HPC Center: ACK Cyfronet AGH) for providing computer facilities and support within computational grant no. PLG/2023/016302. Some experiments were performed on servers purchased with funds from the Priority Research Area (Artificial Intelligence Computing Center Core Facility) under the Strategic Programme Excellence Initiative at Jagiellonian University.

\bibliography{arxiv}
\bibliographystyle{arxiv}

\newpage
\appendix
\onecolumn

\section{SVD.}
In Theorem~\ref{theorem:svd}, we present the formal definition of Singular Value Decomposition from \citep{horn2012matrix}.

\begin{theorem}
\label{theorem:svd}
Let consider a linear operator $\mathbf{A} \in \mathcal{M}_{n,m}$, let $q = min\{m, n\}$, and suppose that $R$ is a rank of $\mathbf{A}$.

\begin{enumerate}
    \item There are unitary matrices $\mathbf{U} \in \mathcal{M}_n$ and $\mathbf{V} \in \mathcal{M}_m$, and a square diagonal matrix \\
    \begin{equation}
    \mathbf{\Sigma}_q = 
    \left[\begin{smallmatrix}
    \sigma_{1} & & 0 \\
    & \ddots & \\
   0 & & \sigma_{q}
   \end{smallmatrix}\right]
    \end{equation}
    
    such that $\sigma_1\geq\sigma_2\geq\cdots\geq\sigma_R\geq0=\sigma_{R+1}=\cdots=\sigma_q$ and 
    \begin{equation}
    \label{eq:svd_original}
        \mathbf{A} = \mathbf{U \Sigma V^{*}}
    \end{equation} in which: $\mathbf{\Sigma} = \mathbf{\Sigma_q}$ if~$m=n$;  $\mathbf{\Sigma} = \begin{bmatrix}\mathbf{\Sigma_q} & 0 \end{bmatrix} \in \mathcal{M}_{n,m}$ if $m>n$; and $\mathbf{\Sigma} = \begin{bmatrix}\mathbf{\Sigma_q} \\ 0 \end{bmatrix} \in \mathcal{M}_{n,m}$ if $m<n$. \\
    \item The parameters $\sigma_1, \dots, \sigma_R$ are the positive square roots of the decreasingly ordered nonzero eigenvalues of $\mathbf{AA^{*}}$, which are the same as the decreasingly nonzero eigenvalues of $\mathbf{A^{*}A}$.
\end{enumerate}
\end{theorem}

\section{Proof of Theorem 4.1.}
\textit{Proof.} Observe that every element of $S^r_{A, B}$ is of rank at most $r$. By the matrix approximation lemma, known as the \textit{Eckart–Young–Mirsky theorem}, the optimal approximation of a matrix $G$ in the Frobenius norm among the rank-$r$ matrices is given by $U_r$, $\Sigma_r$,$V_r^T$, where the decomposition $\mathbf{G} = \mathbf{U}\mathbf{\Sigma}\mathbf{V^T}$ is the SVD of $G$. $\qedsymbol$

\section{Unlearning losses.}

To finetune the set of parameters $\{R_i\}_{i=1}^{L}$ for unlearning, we follow the random labeling unlearning losses proposed by SalUn. Specifically, we use the classification loss $L_c$:
\begin{equation}
\label{eq:classification_loss}
    \min_{\Delta\theta} ~ L_c (\theta_\mathrm{u}) := \mathbb E_{(\mathbf x, y) \sim \mathcal{D}_\mathrm{f}, y^\prime \neq y} \left [ \ell_\mathrm{CE}(\theta_\mathrm{u}; \mathbf x, y^\prime) \right ] 
    + \alpha \mathbb E_{(\mathbf x, y) \sim \mathcal{D}_\mathrm{r}} \left [ \ell_\mathrm{CE}(\theta_\mathrm{u}; \mathbf x, y) \right ],
\end{equation}
where $\ell_{\mathrm{CE}}$ denotes the cross-entropy loss, and $\alpha$ controls the contribution of the remaining dataset $\mathcal{D}_r$.

For the generation task, we apply the generation loss $L_g$:
\begin{equation}
\label{eq:generation_loss}
    \min_{\Delta\theta} ~  L_g (\theta_\mathrm{u}) :=  \mathbb{E}_{(\mathbf x, c) \sim \mathcal D_\mathrm{f}, t, \epsilon \sim \mathcal{N}(0,1), c^\prime \neq c  } \left [ \| \epsilon_{\theta_\mathrm{u}}(\mathbf x_t | c^\prime) - \epsilon_{\theta_\mathrm{u}}(\mathbf x_t | c) \|_2^2 \right ] + \beta \ell_\mathrm{MSE}(\theta_\mathrm{u}; \mathcal D_\mathrm{r}),
\end{equation}
where $\ell_\mathrm{MSE}$ is the mean squared error loss, and $\beta$ controls the contribution of $\mathcal{D}_r$.

Moreover, our method effectively handles both situations, i.e. with access to the remaining dataset $\mathcal{D}_r$ and without it. In the case where the remaining dataset is unavailable, the situation is equivalent to setting $\alpha=0$ and $\beta=0$.

\newpage
\section{Algorithms for \our{} unlearning.}
\label{sec:algorithms}

In this, we present the pseudo-codes for each algorithm used for \our{}. The Section is structured as follows -- firstly, we present the general procedure for selecting weights in \our{}, Alg.~\ref{alg:semu_weights}. Then, we introduce using \our{} in a classification setting~(\ref{alg:semu_classification}), and in the image generation one~(\ref{alg: semu_generation}).   

\begin{algorithm}[H]
\caption{Pseudo code of \our{} selecting weights procedure.}\label{alg:semu_weights}

\begin{algorithmic}

\Require Forgetting set $\mathcal{D}_f$, original model $\mathrm{\theta_o}$, explanation parameter $\gamma$, and forgetting loss function $\ell$.

\Procedure{\our{}\_weights\_selection}{$\mathcal{D}_f$, $\mathrm{\theta_o}$, $\gamma$, $\ell$}

\State $\mathbf{g} \gets \varnothing$ \Comment{Array of accumulated gradients corresponding to $\mathrm{\theta_o}$}

\For{$\mathbf b \gets$ all batches of $\mathcal{D}_f$}
    \State $\mathbf{g} \gets \mathbf{g} + \nabla_\mathrm{\theta_o} \left(- \ell (\mathrm{\theta_o}; \mathbf b) \right)$
    \Comment{Accumulating gradients according to Eq.~\ref{eq:gradient_selection}}
\EndFor

\color{red}\State $\mathbf{\theta_{c}} \gets \varnothing$ \Comment{Original model with changed layers}
\State $\mathbf{\theta_{u}} \gets \varnothing$ \Comment{Trainable parameters for unlearning procedure}
\color{black}
\For{$layer \gets 1 \ldots L$}
    \State $g_l \gets g |_{\mathbf{\theta_l} \subseteq \mathbf{\theta_o}}$ \Comment{Selecting gradients corresponding to $l$\textit{-th} layer}
    \State $g_{l\perp{\mathbf{\theta_l}}} \gets g_l - \frac{\langle g_l, \mathbf{\theta_l} \rangle}{\|\mathbf{\theta_l}\|^2}\mathbf{\theta_l}$ \Comment{Perpendicular projection onto the $l$\textit{-th} layer weights space $\mathbf{\theta_l}$ from Eq.~\ref{eq:projection}}
    \State $U_l$, $\Sigma_l$, $V_l$ $\gets$ \Call{SVD}{ $ \left( g_{l\perp{\mathbf{\theta_l}}} \right) $} \Comment{\textbf{SVD} projection on the $g_{l\perp{\mathbf{\theta_l}}}$ via Eq.~\ref{eq:svd_gradient}}
    \State $r_l \gets \argmin_k e_k \geq \gamma$ \Comment{Selecting the low-rank $r_l$ in the SVD projection with Eq.~\ref{eq:selecting_r}}
    \State $ R_{l, r_l} \gets \mathbf{0}$  \Comment{Initializing the trainable parameters}
    \State $\mathbf{\theta_l} \gets \mathbf{\theta_l} + U_{l, r_l} R_{l, r_l} V_{l, r_l}^T$ \Comment{Updating the $l$\textit{-th} layer parameters with truncated SVD matrices (Eq.~\ref{eq:changed_layer})}
    \color{red}\State $\mathbf{\theta_{c}} \gets \mathbf{\theta_{c}} \cup \mathbf{\theta_l}$ \Comment{Updating original model}
    \State $\mathbf{\theta_{u}} \gets \mathbf{\theta_{u}} \cup R_{l, r_l}$ \Comment{Updating the set of trainable parameters}
    \color{black}
\EndFor
\State \Return $\mathbf{\theta_{c}}$, $\mathbf{\theta_{u}}$
\EndProcedure
\end{algorithmic}
\end{algorithm}

\begin{algorithm}
\caption{Pseudo code of {\our{}} in classification tasks.}\label{alg:semu_classification}
\begin{algorithmic}

\State \hspace{-3.45mm}\textbf{Hyper-parameters:} learning rate $\eta$, explanation parameter $\gamma$, forgetting loss function $\ell$, and number of epochs $E$.

\Require Relabeled forgetting set $\mathcal{D}_f' = \{(\mathbf x_i, c') | (\mathbf x_i, c_i) \in \mathcal{D}_f, c' \neq c_i\}$

\State $\theta_\mathrm{o}, \theta_\mathrm{u}$ $\gets$ \Call{\our{}\_weights\_selection}{$\mathcal{D}_f$, $\theta_\mathrm{o}$, $\gamma$, $\ell$} \Comment{Updating $\theta_\mathrm{o}$ and setting trainable parameters $\theta_\mathrm{u}$ with Alg.~\ref{alg:semu_weights}}

\State $\mathcal D' \gets \mathcal{D}_f' \cup \varnothing$  \color{red} $\left( \mathcal D' \gets \mathcal{D}_f' \cup \mathcal{D}_r \right)$  \Comment{When using \textit{retrain} mode}

\color{black}
\For{$epoch \gets 0 \ldots E-1$}
    \For{$\mathbf b \gets$ all batches of $\mathcal D'$}
    \State $\mathbf{g} \gets \nabla_\mathrm{\theta} L_c (\mathrm{\theta}; \mathbf b)|_{\mathrm{\theta}=\theta_{\mathrm{u}}}$
    \Comment{Batch-wise loss from Eq.~\ref{eq:classification_loss}}
    \State $\theta_{\mathrm{u}} \gets \theta_{\mathrm{u}} - \eta  \mathbf{g}$
    \Comment{One step SGD}
    \EndFor
\EndFor\\
\Return $\theta_u$
\end{algorithmic}
\end{algorithm}

\begin{algorithm}
\caption{Pseudo code of {\our{}} in generation tasks.}\label{alg: semu_generation}
\begin{algorithmic}

\State \hspace{-3.45mm}\textbf{Hyper-parameters:} learning rate $\eta$, explanation parameter $\gamma$, forgetting loss function $\ell$, and number of iterations $T$.

\Require Relabeled forgetting set $\mathcal{D}_f' = \{(\mathbf x_i, c') | (\mathbf x_i, c_i) \in \mathcal{D}_f, c' \neq c_i\}$

\State $\theta_\mathrm{o}, \theta_\mathrm{u}$ $\gets$ \Call{\our{}\_weights\_selection}{$\mathcal{D}_f$, $\theta_\mathrm{o}$, $\gamma$, $\ell$} \Comment{Updating $\theta_\mathrm{o}$ and setting trainable parameters $\theta_\mathrm{u}$ with Alg.~\ref{alg:semu_weights}}

\State $\mathcal D' \gets \mathcal{D}_f' \cup \varnothing$  \color{red} $\left( \mathcal D' \gets \mathcal{D}_f' \cup \mathcal{D}_r \right)$  \Comment{When using \textit{retrain} mode}

\color{black}
\For{$it \gets 0 \ldots T-1$}
    \State Sampling batch $\mathbf b$ from $\mathcal D'$
   \State $\mathbf{g} \gets \nabla_\mathrm{\theta} L_g (\mathrm{\theta}; \mathbf b)|_{\mathrm{\theta}=\theta_{\mathrm{u}}}$
    \Comment{Batch-wise loss from \eqref{eq:generation_loss}}
    \State $\theta_{\mathrm{u}} \gets \theta_{\mathrm{u}} - \eta  \mathbf{g}$ \Comment{One step SGD}
\EndFor
\State \Return $\theta_u$
\end{algorithmic}
\end{algorithm}

\newpage

\section{Additional results for SEMU in image classification task.}

\begin{table*}[htbp]\small
\centering
\caption{Comparison of methods for Random Data Forgetting (10\% and 50\%) on ResNet-18 with CIFAR-10. The table reports Unlearning Accuracy (UA), Remaining Accuracy (RA), Testing Accuracy (TA), and Membership Inference Attack (MIA), with values in parentheses showing differences from the \textit{Retrain} baseline. TParams denotes the percentage of trained parameters relative to standard ResNet-18 (not unforgetting). Note that we bold results achieving the closest TA accuracy to Retrain and those which alter the smallest portion of model's weigths.}
\label{tab:resnet18_cifar10}
\resizebox{\textwidth}{!}{
\begin{tabular}{@{}l@{}c@{\;\;}c@{\;\;}c@{\;\;}c@{\;\;}c@{\quad}c@{\;\;}c@{\;\;}c@{\;\;}c@{\;\;}c@{}}
\toprule
\multirow{2}{*}{Methods} & \multicolumn{5}{c}{Random Data Forgetting (10\%)} & \multicolumn{5}{c}{Random Data Forgetting (50\%)} \\
\cmidrule(lr){2-6} \cmidrule(lr){7-11}
 & UA & RA & TA & MIA & TParams & UA & RA & TA & MIA & TParams \\
\midrule
Retrain & $5.24$ \textcolor{blue}{(0.00)} & $100.00$ \textcolor{blue}{(0.00)} & $94.26$ \textcolor{blue}{(0.00)} & $12.88$ \textcolor{blue}{(0.00)} & 100\% & $7.91$  \textcolor{blue}{(0.00)} & $100.00$ \textcolor{blue}{(0.00)} & $91.72$ \textcolor{blue}{(0.00)} & $19.29$ \textcolor{blue}{(0.00)} & 100\% \\
\cmidrule(lr){1-11}
FT & $0.63$ \textcolor{blue}{(4.61)} & $99.88$ \textcolor{blue}{(0.12)} & $94.06$ \textcolor{blue}{(0.20)} & $2.70$ \textcolor{blue}{(10.19)} & 100\% & $0.44$ \textcolor{blue}{(7.47)} & $99.96$ \textcolor{blue}{(0.04)} & $94.23$ \textcolor{blue}{(2.52)} & $2.15$ \textcolor{blue}{(17.14)} & 100\% \\
RL & $7.61$ \textcolor{blue}{(2.37)} & $99.67$ \textcolor{blue}{(0.33)} & $92.83$ \textcolor{blue}{(1.43)} & $37.36$ \textcolor{blue}{(24.47)} & 100\% & $4.80$ \textcolor{blue}{(3.11)} & $99.55$ \textcolor{blue}{(0.45)} & $91.31$ \textcolor{blue}{(0.40)} & $41.95$ \textcolor{blue}{(22.66)} & 100\% \\
GA & $0.69$ \textcolor{blue}{(4.56)} & $99.50$ \textcolor{blue}{(0.50)} & $94.01$ \textcolor{blue}{(0.25)} & $1.70$ \textcolor{blue}{(11.18)} & 100\% & $0.40$ \textcolor{blue}{(7.50)} & $99.61$ \textcolor{blue}{(0.39)} & $94.34$ \textcolor{blue}{(2.63)} & $1.22$ \textcolor{blue}{(18.07)} & 100\% \\
IU & $1.07$ \textcolor{blue}{(4.17)} & $99.20$ \textcolor{blue}{(0.80)} & $93.20$ \textcolor{blue}{(1.06)} & $2.67$ \textcolor{blue}{(10.21)} & 100\% & $3.97$ \textcolor{blue}{(3.94)} & $96.21$ \textcolor{blue}{(3.79)} & $90.00$ \textcolor{blue}{(1.71)} & $7.29$ \textcolor{blue}{(12.00)} & 100\% \\
BE & $0.59$ \textcolor{blue}{(4.65)} & $99.42$ \textcolor{blue}{(0.58)} & $93.85$ \textcolor{blue}{(0.42)} & $7.47$ \textcolor{blue}{(5.41)} & 100\% & $3.08$ \textcolor{blue}{(4.82)} & $96.84$ \textcolor{blue}{(3.16)} & $90.41$ \textcolor{blue}{(1.31)} & $24.87$ \textcolor{blue}{(5.58)} & 100\% \\
BS & $1.78$ \textcolor{blue}{(3.47)} & $98.29$ \textcolor{blue}{(1.71)} & $92.69$ \textcolor{blue}{(1.57)} & $8.96$ \textcolor{blue}{(3.93)} & 100\% & $9.76$ \textcolor{blue}{(1.85)} & $90.19$ \textcolor{blue}{(9.81)} & $83.71$ \textcolor{blue}{(8.01)} & $32.15$ \textcolor{blue}{(12.86)} & 100\% \\
$\ell_1$-sparse & $4.19$ \textcolor{blue}{(1.06)} & $97.74$ \textcolor{blue}{(2.26)} & $91.59$ \textcolor{blue}{(2.67)} & $9.84$ \textcolor{blue}{(3.04)} & 100\% & $1.44$ \textcolor{blue}{(6.47)} & $99.52$ \textcolor{blue}{(0.48)} & $93.13$ \textcolor{blue}{(1.41)} & $4.76$ \textcolor{blue}{(14.52)} & 100\% \\
SalUn & $2.85$ \textcolor{blue}{(2.39)} & $99.62$ \textcolor{blue}{(0.38)} & $93.93$ \textcolor{blue}{(0.33)} & $14.39$ \textcolor{blue}{(1.51)} & 100\% & $7.75$ \textcolor{blue}{(0.16)} & $94.28$ \textcolor{blue}{(5.72)} & $89.29$ \textcolor{blue}{(2.43)} & $16.99$ \textcolor{blue}{(2.30)} & 100\% \\
SalUn-soft & $4.19$ \textcolor{blue}{(1.06)} & $99.74$ \textcolor{blue}{(0.26)} & $93.44$ \textcolor{blue}{(0.83)} & $19.49$ \textcolor{blue}{(6.61)} & 100\% & $3.41$ \textcolor{blue}{(4.49)} & $99.62$ \textcolor{blue}{(0.38)} & $91.82$ \textcolor{blue}{(0.11)} & $31.50$ \textcolor{blue}{(12.21)} & 100\% \\
\cmidrule(lr){1-11}
\our{} & $0.60$ \textcolor{blue}{(4.64)} & $99.40$ \textcolor{blue}{(0.60)} & $\mathbf{94.22}$ \textbf{\textcolor{blue}{(0.04)}} & $5.40$ \textcolor{blue}{(7.48)} & \textbf{0.54}\%
& $1.77$ \textcolor{blue}{(6.14)} & $98.12$ \textcolor{blue}{(1.88)} & $91.80$ \textcolor{blue}{(0.08)} & $7.20$ \textcolor{blue}{(12.09)} & \textbf{0.64}\% \\ 
\our{}$_{remain}$  & $0.69$ \textcolor{blue}{(4.55)} & $99.43$ \textcolor{blue}{(0.57)} & $\mathbf{94.30}$ \textbf{\textcolor{blue}{(0.04)}} & $5.51$ \textcolor{blue}{(7.37)} & \textbf{0.54\%}
& $1.82$ \textcolor{blue}{(6.09)} & $98.12$ \textcolor{blue}{(1.88)} & $\mathbf{91.72}$ \textbf{\textcolor{blue}{(0.00)}} & $7.54$ \textcolor{blue}{(11.75)} & 0.72\% \\
\bottomrule
\end{tabular}
}
\end{table*}

\begin{table*}[htbp]\small
\centering
\caption{Comparison of methods for Random Data Forgetting (10\% and 50\%) on VGG-16 with CIFAR-10 dataset. The table reports Unlearning Accuracy (UA), Remaining Accuracy (RA), Testing Accuracy (TA), and Membership Inference Attack (MIA), with values in parentheses showing differences from the \textit{Retrain} baseline. TParams denotes the percentage of trained parameters relative to standard VGG-16 (not unforgetting). Note that we bold results achieving the closest TA accuracy to Retrain and those which alter the smallest portion of model's weigths.}
\label{tab:vgg16_cifar10}
\resizebox{\textwidth}{!}{
\begin{tabular}{@{}l@{}c@{\;\;}c@{\;\;}c@{\;\;}c@{\;\;}c@{\quad}c@{\;\;}c@{\;\;}c@{\;\;}c@{\;\;}c@{}}
\toprule
\multirow{2}{*}{Methods} & \multicolumn{5}{c}{Random Data Forgetting (10\%)} & \multicolumn{5}{c}{Random Data Forgetting (50\%)} \\
\cmidrule(lr){2-6} \cmidrule(lr){7-11}
 & UA & RA & TA & MIA & TParams & UA & RA & TA & MIA & TParams \\
\midrule
Retrain & $5.98$ & $99.99$ & $93.06$ & $10.36$ & 100\% & $9.47$ & $100.00$ & $90.18$ & $16.64$ & 100\% \\
\cmidrule(lr){1-11}
FT & $1.51$ \textcolor{blue}{(4.47)} & $99.54$ \textcolor{blue}{(0.45)} & $92.64$ \textcolor{blue}{(0.42)} & $3.76$ \textcolor{blue}{(6.60)} & 100\% & $5.70$ \textcolor{blue}{(3.77)} & $97.51$ \textcolor{blue}{(2.49)} & $89.37$ \textcolor{blue}{(0.81)} & $12.20$ \textcolor{blue}{(4.44)} & 100\% \\
RL & $5.71$ \textcolor{blue}{(0.27)} & $99.65$ \textcolor{blue}{(0.34)} & $92.29$ \textcolor{blue}{(0.77)} & $15.98$ \textcolor{blue}{(5.62)} & 100\% & $4.09$ \textcolor{blue}{(5.38)} & $96.77$ \textcolor{blue}{(3.23)} & $89.91$ \textcolor{blue}{(0.27)} & $13.88$ \textcolor{blue}{(2.76)} & 100\% \\
GA & $0.93$ \textcolor{blue}{(5.05)} & $99.37$ \textcolor{blue}{(0.62)} & $93.63$ \textcolor{blue}{(0.57)} & $1.36$ \textcolor{blue}{(9.00)} & 100\% & $0.63$ \textcolor{blue}{(8.84)} & $99.38$ \textcolor{blue}{(0.62)} & $93.64$ \textcolor{blue}{(3.46)} & $1.15$ \textcolor{blue}{(15.49)} & 100\% \\
IU & $1.69$ \textcolor{blue}{(4.29)} & $98.78$ \textcolor{blue}{(1.21)} & $91.69$ \textcolor{blue}{(1.37)} & $2.71$ \textcolor{blue}{(7.65)} & 100\% & $5.71$ \textcolor{blue}{(3.76)} & $94.56$ \textcolor{blue}{(5.44)} & $87.23$ \textcolor{blue}{(2.95)} & $8.34$ \textcolor{blue}{(8.30)} & 100\% \\
BE & $0.80$ \textcolor{blue}{(5.18)} & $99.39$ \textcolor{blue}{(0.60)} & $93.68$ \textcolor{blue}{(0.62)} & $1.42$ \textcolor{blue}{(8.94)} & 100\% & $20.58$ \textcolor{blue}{(11.11)} & $79.40$ \textcolor{blue}{(20.60)} & $72.58$ \textcolor{blue}{(17.60)} & $11.74$ \textcolor{blue}{(4.90)} & 100\% \\
BS & $0.80$ \textcolor{blue}{(5.18)} & $99.40$ \textcolor{blue}{(0.59)} & $93.68$ \textcolor{blue}{(0.62)} & $1.38$ \textcolor{blue}{(8.98)} & 100\% & $2.44$ \textcolor{blue}{(7.03)} & $97.56$ \textcolor{blue}{(2.44)} & $89.69$ \textcolor{blue}{(0.49)} & $4.90$ \textcolor{blue}{(11.74)} & 100\% \\
$\ell_1$-sparse & $4.98$ \textcolor{blue}{(1.00)} & $97.03$ \textcolor{blue}{(2.96)} & $90.15$ \textcolor{blue}{(2.91)} & $9.69$ \textcolor{blue}{(0.67)} & 100\% & $3.13$ \textcolor{blue}{(6.34)} & $98.77$ \textcolor{blue}{(1.23)} & $91.01$ \textcolor{blue}{(0.83)} & $7.06$ \textcolor{blue}{(9.58)} & 100\% \\
SalUn & $3.89$ \textcolor{blue}{(2.09)} & $98.74$ \textcolor{blue}{(1.25)} & $91.62$ \textcolor{blue}{(1.44)} & $9.96$ \textcolor{blue}{(0.40)} & 100\% & $3.02$ \textcolor{blue}{(6.45)} & $98.14$ \textcolor{blue}{(1.86)} & $89.82$ \textcolor{blue}{(0.36)} & $15.15$ \textcolor{blue}{(1.49)} & 100\% \\
SalUn-soft & $5.24$ \textcolor{blue}{(0.74)} & $99.70$ \textcolor{blue}{(0.29)} & $92.26$ \textcolor{blue}{(0.80)} & $12.31$ \textcolor{blue}{(1.95)} & 100\% & $3.44$ \textcolor{blue}{(6.03)} & $99.64$ \textcolor{blue}{(0.36)} & $91.11$ \textcolor{blue}{(0.93)} & $16.19$ \textcolor{blue}{(0.45)} & 100\% \\
\cmidrule(lr){1-11}
\our{} & $0.67$ \textcolor{blue}{(5.31)} & $99.33$ \textcolor{blue}{(0.66)} & $\mathbf{93.09}$ \textbf{\textcolor{blue}{(0.03)}} & $5.02$ \textcolor{blue}{(5.34)} & 0.89\% 
& $3.31$ \textcolor{blue}{(6.16)} & $96.32$ \textcolor{blue}{(3.68)} & $90.01$ \textcolor{blue}{(0.17)} & $18.92$ \textcolor{blue}{(2.28)} & 0.34\% \\ 
\our{}$_{remain}$  & $0.62$ \textcolor{blue}{(5.36)} & $99.37$ \textcolor{blue}{(0.62)} & $93.27$ \textcolor{blue}{(0.21)} & $7.02$ \textcolor{blue}{(3.34)} & \textbf{0.23\%} 
& $2.56$ \textcolor{blue}{(6.91)} & $96.98$ \textcolor{blue}{(3.02)} & $\mathbf{90.21}$ \textbf{\textcolor{blue}{(0.03)}} & $16.06$ \textcolor{blue}{(0.58)} & \textbf{0.29\%} \\ 
\bottomrule
\end{tabular}
}
\end{table*}

\begin{table*}[htbp]\small
\centering
\caption{Performance evaluation for class-wise forgetting on ResNet-18, pre-trained on the CIFAR-10 dataset. The table presents the results of various methods in terms of Unlearning Accuracy (UA), Remaining Accuracy (RA), Testing Accuracy (TA), and Membership Inference Attack (MIA). The values in parentheses indicate the difference compared to the \textit{Retrain} baseline.}
\label{tab:classwise_forgetting}
\begin{tabular}{lccccc}
\toprule
Methods & UA & RA & TA & MIA & TParams \\ 
\midrule
Retrain          & 100.0      & 100.0      & 92.47       & 100.0     & 100\%    \\
\cmidrule(lr){1-6}
FT               & 31.69 \textcolor{blue}{(68.31)} & 99.92 \textcolor{blue}{(0.07)} & 94.78 \textcolor{blue}{(2.31)} & 93.53 \textcolor{blue}{(6.47)}  & 100\% \\
RL               & 89.33 \textcolor{blue}{(10.67)} & 99.92 \textcolor{blue}{(0.08)} & 94.52 \textcolor{blue}{(2.06)} & 100.0 \textcolor{blue}{(0.00)}  & 100\% \\
GA               & 99.91 \textcolor{blue}{(0.09)}  & 38.92 \textcolor{blue}{(61.07)} & 38.18 \textcolor{blue}{(54.29)} & 99.98 \textcolor{blue}{(0.02)}   & 100\% \\
IU               & 97.02 \textcolor{blue}{(2.98)}  & 94.78 \textcolor{blue}{(5.22)}  & 89.10 \textcolor{blue}{(3.37)}  & 99.13 \textcolor{blue}{(0.87)}   & 100\% \\
BE               & 79.13 \textcolor{blue}{(20.87)} & 97.71 \textcolor{blue}{(2.29)}  & 91.88 \textcolor{blue}{(0.59)}  & 93.60 \textcolor{blue}{(6.40)}   & 100\% \\
BS               & 79.60 \textcolor{blue}{(20.40)} & 97.79 \textcolor{blue}{(2.21)}  & \textbf{91.94 \textcolor{blue}{(0.52)}}  & 93.42 \textcolor{blue}{(6.58)}  & 100\%  \\
$\ell_1$-sparse  & 100.0 \textcolor{blue}{(0.00)} & 97.92 \textcolor{blue}{(2.08)}  & 92.29 \textcolor{blue}{(0.18)}  & 100.0 \textcolor{blue}{(0.00)}  & 100\% \\
SalUn            & 99.91 \textcolor{blue}{(0.09)}  & 99.93 \textcolor{blue}{(0.07)}  & 94.56 \textcolor{blue}{(2.09)}  & 100.0 \textcolor{blue}{(0.00)}  & 50\% \\
SalUn-soft       & 97.13 \textcolor{blue}{(2.87)}  & 99.88 \textcolor{blue}{(0.12)}  & 94.64 \textcolor{blue}{(2.18)}  & 100.0 \textcolor{blue}{(0.00)} & 50\% \\ 
\midrule
\our{} & $99.83$ \textcolor{blue}{(0.17)} & $98.22$ \textcolor{blue}{(1.78)} & $92.26$ \textcolor{blue}{(0.60)} & $100.00$ \textcolor{blue}{(0.00)} & 0.87\% \\
\our{}$_{remain}$ & $99.99$ \textcolor{blue}{(0.01)} & $99.48$ \textcolor{blue}{(0.52)} & $94.76$ \textcolor{blue}{(2.29)} & $100.00$ \textcolor{blue}{(0.00)} & \textbf{0.63\%} \\
\bottomrule
\end{tabular}
\end{table*}

\begin{figure}[htbp]
    \centering
    \includegraphics[width=\textwidth]{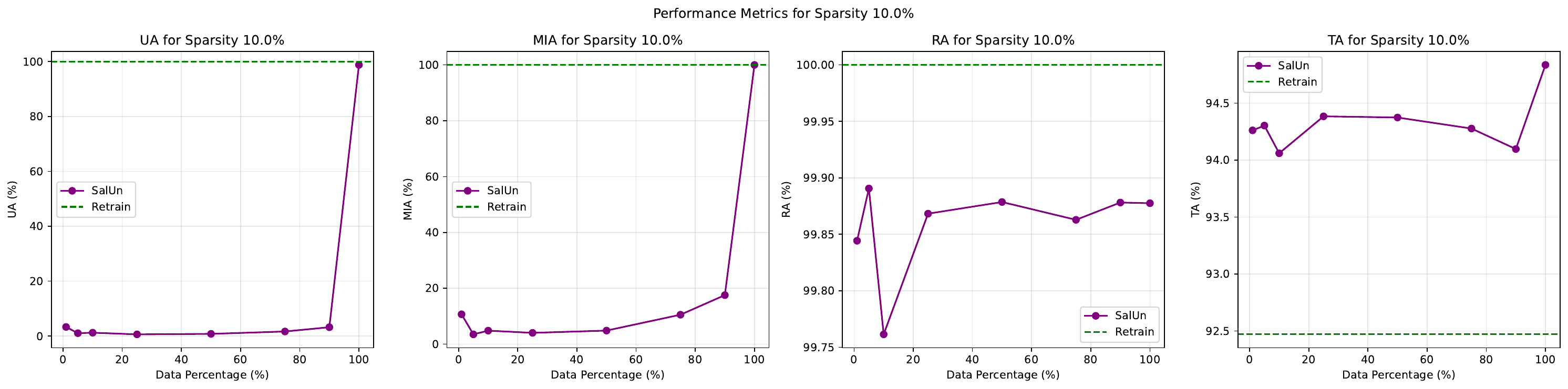}
    \includegraphics[width=\textwidth]{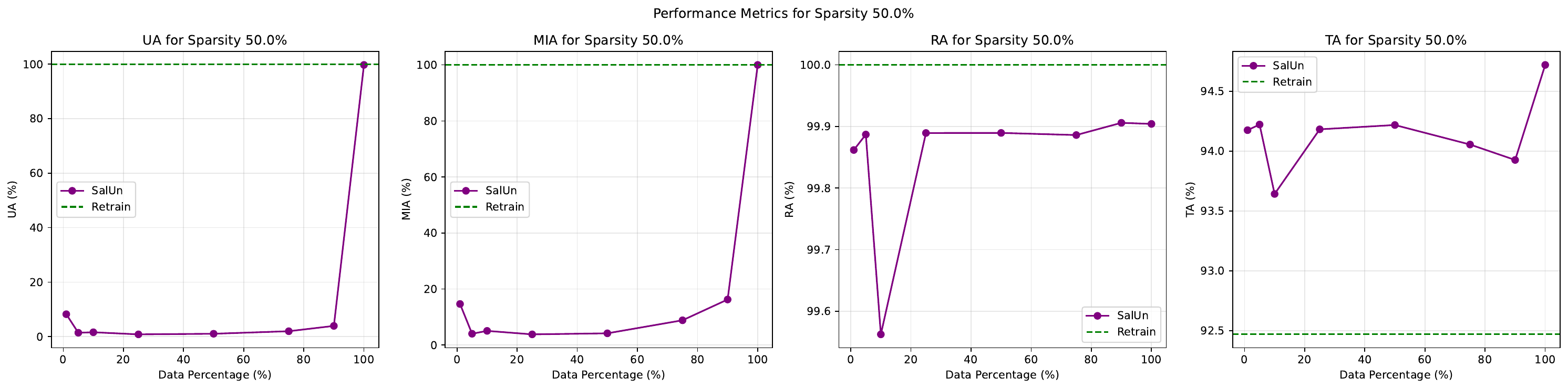}
    \caption{Overview of SalUn results for the Class-Wise Forgetting scenario on ResNet-18 with CIFAR-10 for different percentages of available data from the class selected to forget. The top row depicts results for $10\%$, while the bottom row shows scores for $50\%$ of the saliency sparsity. The plots in consecutive columns demonstrate Unlearning Accuracy (UA), Membership Inference Attack (MIA), Remaining Accuracy (RA) and Testing Accuracy (TA), respectively. In all cases, results are compared to the \textit{Retrain} baseline.}
    \label{fig:ablation_class_wise_01}
\end{figure}

\begin{table*}[htbp]\small
\centering
\caption{Comparison of SalUn results for Class-Wise Forgetting on ResNet-18 with CIFAR-10 for different percentages of available data from the class selected to forget (from $1$ to $100\%$), and different saliency sparsity ($10$ and $50\%$). The table reports Unlearning Accuracy (UA), Remaining Accuracy (RA), Testing Accuracy (TA), and Membership Inference Attack (MIA), with values in parentheses showing differences from the \textit{Retrain} baseline.}
\label{tab:resnet18_cifar10_ablation}

\resizebox{\textwidth}{!}{
\begin{tabular}{@{}l@{}c@{\;\;}c@{\;\;}c@{\;\;}c@{\;\;}c@{\quad}c@{\;\;}c@{\;\;}c@{\;\;}c@{\;\;}c@{}}
\toprule
\multirow{2}{*}{Available data} & \multicolumn{4}{c}{Saliency sparsity (10\%)} & \multicolumn{4}{c}{Saliency Sparsity (50\%)} \\
\cmidrule(lr){2-5} \cmidrule(lr){6-9}
 & UA & RA & TA & MIA & UA & RA & TA & MIA  \\
\midrule
Retrain & $100.00$ & $100.00$ & $92.47$ & $100.00$ & $100.00$ & $100.00$ & $92.47$ & $100.00$ \\
\midrule
$1\%$ & $3.33$ \textcolor{blue}{(96.67)} & $99.84$ \textcolor{blue}{(0.16)} & $94.26$ \textcolor{blue}{(1.79)} & $10.67$ \textcolor{blue}{(89.33)} & $8.22$ \textcolor{blue}{(91.78)} & $99.86$ \textcolor{blue}{(0.14)} & $94.18$ \textcolor{blue}{(1.71)} & $14.67$ \textcolor{blue}{(85.33)} \\
$5\%$ & $1.07$ \textcolor{blue}{(98.93)} & $99.89$ \textcolor{blue}{(0.11)} & $94.30$ \textcolor{blue}{(1.83)} & $3.47$ \textcolor{blue}{(96.53)} & $1.42$ \textcolor{blue}{(98.58)} & $99.89$ \textcolor{blue}{(0.11)} & $94.22$ \textcolor{blue}{(1.75)} & $4.04$ \textcolor{blue}{(95.96)} \\
$10\%$ & $1.24$ \textcolor{blue}{(98.76)} & $99.76$ \textcolor{blue}{(0.24)} & $94.06$ \textcolor{blue}{(1.59)} & $4.80$ \textcolor{blue}{(95.20)} & $1.60$ \textcolor{blue}{(98.40)} & $99.56$ \textcolor{blue}{(0.44)} & $93.64$ \textcolor{blue}{(1.17)} & $5.07$ \textcolor{blue}{(94.93)} \\
$25\%$ & $0.64$ \textcolor{blue}{(99.36)} & $99.87$ \textcolor{blue}{(0.13)} & $94.38$ \textcolor{blue}{(1.91)} & $4.01$ \textcolor{blue}{(95.99)} & $0.83$ \textcolor{blue}{(99.17)} & $99.89$ \textcolor{blue}{(0.11)} & $94.18$ \textcolor{blue}{(1.71)} & $3.82$ \textcolor{blue}{(96.18)} \\
$50\%$ & $0.82$ \textcolor{blue}{(99.18)} & $99.88$ \textcolor{blue}{(0.12)} & $94.37$ \textcolor{blue}{(1.90)} & $4.82$ \textcolor{blue}{(95.18)} & $1.05$ \textcolor{blue}{(98.95)} & $99.89$ \textcolor{blue}{(0.11)} & $94.22$ \textcolor{blue}{(1.75)} & $4.20$ \textcolor{blue}{(95.80)}  \\
$75\%$ & $1.68$ \textcolor{blue}{(98.32)} & $99.86$ \textcolor{blue}{(0.14)} & $94.28$ \textcolor{blue}{(1.81)} & $10.51$ \textcolor{blue}{(89.49)} & $1.98$ \textcolor{blue}{(98.02)} & $99.89$ \textcolor{blue}{(0.11)} & $94.06$ \textcolor{blue}{(1.59)} & $8.85$ \textcolor{blue}{(91.15)}  \\
$90\%$ & $3.24$ \textcolor{blue}{(96.76)} & $99.88$ \textcolor{blue}{(0.12)} & $94.10$ \textcolor{blue}{(1.63)} & $17.52$ \textcolor{blue}{(82.48)} & $3.93$ \textcolor{blue}{(96.07)} & $99.91$ \textcolor{blue}{(0.09)} & $93.93$ \textcolor{blue}{(1.46)} & $16.26$ \textcolor{blue}{(83.74)}  \\
$100\%$ & $98.76$ \textcolor{blue}{(1.24)} & $99.88$ \textcolor{blue}{(0.12)} & $94.84$ \textcolor{blue}{(2.37)} & $100.00$ \textcolor{blue}{(0.00)} & $99.71$ \textcolor{blue}{(0.29)} & $99.90$ \textcolor{blue}{(0.10)} & $94.72$ \textcolor{blue}{(2.25)} & $100.00$ \textcolor{blue}{(0.00)} \\
\bottomrule
\end{tabular}
}
\end{table*}

\newpage

\section{Additional results for SEMU in image generation task.}

\subsection{CIFAR10 generation with DDPM.}
\begin{table}[htbp]
\centering
\caption{Class-wise forgetting on classifier-free guidance DDPM.}
\begin{tabular}{c|c|c|c|c}
\toprule[1pt]
\textbf{Methods} & \textbf{UA} ($\uparrow$) & \textbf{TA} ($\uparrow$)  & \textbf{FID} ($\downarrow$) & \textbf{\#Params} ($\downarrow$)\\
\midrule
Retrain & 100.00 & 100.00 & 11.69 & 100\%\\
\midrule
ESD & \textbf{100.00} & \textbf{--} & 17.37 &  \textbf{--}\\
SalUn & 99.20 & 14.22 & \textbf{11.21}  & 50\%\\
\midrule
\our{} & 95.60 & \textbf{14.87} & 16.93 & \textbf{1.2\%}\\
\our{}$_{subset}$  & 99.40 & 14.71 & 13.93 & 1.5\%\\
\our{}$_{remain}$  & \textbf{100.00} & 14.64 & 14.51 & 1.8\%\\
\bottomrule[1pt]
\end{tabular}
\label{tab:ddpm_class_forgetting}
\end{table}

\subsection{ImageNette generation with Stable Diffusion.}

\begin{table}
\caption{
Performance of class-wise forgetting on  Imagenette using SD. The best unlearning performance for each forgetting class is highlighted in \textbf{bold} for  UA and FID, respectively. For \our{}, we averaged FIDs for the smaller number of classes due to biasing of FID metric. 
}
\centering
\begin{tabular}{c|cc|cc|cc|cc|cc
}
\toprule[1pt]
\multirow{2}{*}{\textbf{Forget. Class}} & \multicolumn{2}{c|}{\our{}$_{remain}$}  & \multicolumn{2}{c|}{\our{}} & \multicolumn{2}{c|}{SalUn}          & \multicolumn{2}{c|}{ESD}  & \multicolumn{2}{c}{FMN}  
\\
& \multicolumn{1}{c|}{UA ($\uparrow$)} & FID ($\downarrow$) &
\multicolumn{1}{c|}{UA ($\uparrow$)} & FID ($\downarrow$) &
\multicolumn{1}{c|}{UA ($\uparrow$)} & FID ($\downarrow$) &
\multicolumn{1}{c|}{UA ($\uparrow$)} & FID ($\downarrow$) & \multicolumn{1}{c|}{UA ($\uparrow$)} & FID ($\downarrow$) 
\\
\midrule
Tench   & 89.00 & 11.40 & 94.00 & 2.31  & \textbf{100.00} & 2.53 & 99.40 & \textbf{1.22} & 42.40 & 1.63
\\
English Springer & 94.00 & 4.14 & 94.00 & 2.27 & \textbf{100.00}  & \textbf{0.79} & 100.00 & 1.02 & 27.20 & 1.75\\
Cassette Player & 98.00 & 1.36 & 92.00 & 26.23 & 99.80 & 0.91 & \textbf{100.00} & 1.84 & 93.80 & \textbf{0.80}\\
Chain Saw & 96.00 & 8.54 & 64.00 & 1.12 & \textbf{100.00} & 1.58 & 96.80 & 1.48 & 48.40 & \textbf{0.94}\\
Church & 85.00 & 14.30 & 70.00 & -- & \textbf{99.60} & \textbf{0.90} & 98.60 & 1.91 & 23.80 & 1.32 \\
French Horn & \textbf{100.00} & \textbf{0.81} & 98.00 & 4.20 & \textbf{100.00} & \textbf{0.94} & 99.80 & 1.08 & 45.00 & 0.99\\
Garbage Truck & 99.00 & 2.51 & 86.00 & -- & \textbf{100.00} & \textbf{0.91} & 100.00 & 2.71 & 41.40 & 0.92\\
Gas Pump & 98.00 & 2.48 & 88.00 & 1.32 & \textbf{100.00} & \textbf{1.05} & 100.00 & 1.99 & 53.60 & 1.30\\
Golf Ball & 95.00 & 5.77 & 84.00 & 1.46 & 98.80  & 1.45 & \textbf{99.60} & \textbf{0.80} & 15.40 & 1.05\\
Parachute & 95.00 & 13.85 & 68.00 & --  & \textbf{100.00} & 1.16 & 99.80 & \textbf{0.91} & 34.40 & 2.33\\
\midrule
Average & 94.90 & 6.52 & 83.80 & 5.56 $^*$ & \textbf{99.82}  & \textbf{1.22} & 99.40 & 1.49 & 42.54 & 1.30\\
\bottomrule[1pt]
\end{tabular}
\label{tab:sd}
\end{table}

\newpage

\section{Samples from the diffusion models unlearned with \our{}.}

In this Section, we present the samples from the DDPM model unlearned a CIFAR10's class (\textit{airplanes}) with \our{}.

\subsection{Samples from DDPM on CIFAR10.}

\begin{figure}[h!]
    \centering
    \includegraphics[width=0.45\textwidth]{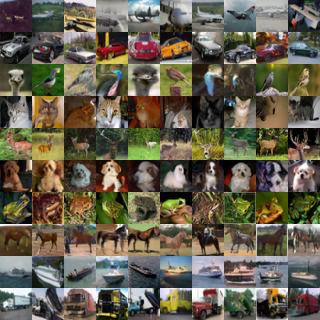}
    \includegraphics[width=0.45\textwidth]{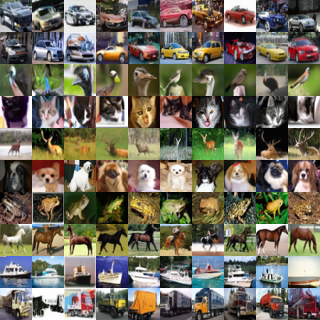}
    \label{fig:samples_ddpm1}
    \caption{Comparison between \our{} with (\textbf{left}) and without (\textbf{right}) access to the remaining dataset. The DDPM model was pretrained on CIFAR10, and with \our{}, we unlearned the class \textit{airplanes} (top rows). We observe that the access to the remaining dataset stabilizes generation and helps to change samples from one class to the other. On the other hand, lack of such an access prevent model from total forgetting.}
\end{figure}

\begin{figure}[h!]
    \centering
    \includegraphics[width=0.45\textwidth]{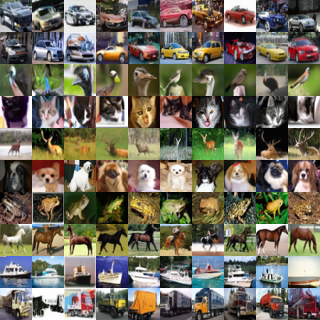}
    \label{fig:samples_ddpm2}
    \caption{Setting in which the unlearned model have an access to the very limited number of samples from the remaining dataset. As we can see, a limited number of additional datapoints is a sufficient for \our{} to have the same quality of samples as the when having access to the whole remaining dataset.}
\end{figure}

\end{document}